\documentclass[journal]{IEEEtran}
\usepackage{graphicx}
\usepackage{indentfirst}
\usepackage{amsmath}
\usepackage{amssymb}
\usepackage{latexsym}
\usepackage{cite}
\usepackage{algorithm}
\usepackage{algorithmic}
\usepackage{multirow}
\usepackage{float} 
\usepackage{subfigure}

\usepackage{cite}

\ifCLASSINFOpdf
\else
\fi
\begin{document}
%
\title{Graph Regularized Nonnegative Tensor Ring
Decomposition for Multiway Representation Learning}
%
%
%

\author{Yuyuan~Yu, Guoxu~Zhou, Ning~Zheng,                Shengli~Xie,~\IEEEmembership{Fellow,~IEEE}
        and~Qibin~Zhao
\thanks{Yuyuan Yu is with the School of Automation, Guangdong University of
	Technology, Guangzhou 510006, China (e-mail: isyuyuyuan@hotmail.com).}
\thanks{Guoxu Zhou is with the School of Automation, Guangdong University of Technology, Guangzhou 510006, China, and also with the Key Laboratory of Intelligent Detection and The Internet of Things in Manufacturing, Ministry of Education, Guangzhou 510006, China (e-mail: gx.zhou@gdut.edu.cn.}
\thanks{Ning Zheng is with the Research Center for Statistical Machine Learning, 	The Institute of Statistical Mathematics, 10-3 Midori-cho, Tachikawa Tokyo 190-8562, Japan (e-mail: nzheng@ism.ac.jp)}
\thanks{Shengli Xie is with the School of Automation, Guangdong University
	of Technology, Guangzhou 510006, China, and also with the Guangdong– Hong Kong–Macao Joint Laboratory for Smart Discrete Manufacturing, Guangdong University of Technology, Guangzhou 510006, China (e-mail: shlxie@gdut.edu.cn).}
\thanks{Qibin Zhao is with the School of Automation, Guangdong University of
	Technology, Guangzhou 510006, China, also with the Joint International
	Research Laboratory of Intelligent Information Processing and System Integration of IoT, Ministry of Education, Guangdong University of Technology,
	Guangzhou 510006, China, and also with the Center for Advanced Intelligence
	Project (AIP), RIKEN, Tokyo, 103-0027, Japan (e-mail: qibin.zhao@riken.jp)}}
\maketitle

\begin{abstract}
Tensor ring (TR) decomposition is a powerful tool for exploiting the low-rank nature of multiway data and has demonstrated great potential in a variety of important applications. In this paper, nonnegative tensor ring (NTR) decomposition and graph regularized NTR (GNTR) decomposition are proposed, where the former equips TR decomposition with local feature extraction by imposing nonnegativity on the core tensors and the latter is additionally able to capture manifold geometry information of tensor data, both significantly extend the applications of TR decomposition for nonnegative multiway representation learning. Accelerated proximal gradient based methods are derived for NTR and GNTR. The experimental result demonstrate that the proposed algorithms can extract parts-based basis with rich colors and rich lines from tensor objects that provide more interpretable and meaningful representation, and hence yield better performance than the state-of-the-art tensor based methods in clustering and classification tasks.
\end{abstract}


%
\IEEEpeerreviewmaketitle

\section{Introduction}
%
%
%
%
\IEEEPARstart{E}{xtracting} meaningful and interpretable low-dimensional representation from high-dimensional data is a fundamental task in the fields of data mining, signal processing, and machine learning. One of the main challenges is how to capture the low-dimensional features from high-dimensional data with a physical meaning while providing task-related interpretability. Nonnegative matrix factorization (NMF) \cite{lee1999learning} has gained much attention recently, which aims to approximate the data matrix by the mutiplication of  two nonnegative factor matrices. It restricts the components to be nonnegative, which makes the components have sparsity and suppresses the energy of the small amount of noise in the data \cite{gillis2014and}. Particularly, NMF has the ability for learning localized parts of nonnegative tensor objects and can give a physically meaningful and more interpretable result, which has been applied to feature extraction \cite{zhou2015efficient,zhang2012low,wang2015semi}, sparse coding \cite{hoyer2004non}, multi-view clustering \cite{li2019deep,liang2020multi,liang2020semi}, etc.

With the advancement of data acquisition technology, more and more related high-dimensional data are collected, which is also known as the multiway data or tensor, such as color images, video, and so on. Nonnegative tensor decomposition \cite{kim2007nonnegative,hazan2005sparse,zhou2014nonnegative,cichocki2009nonnegative} has also gained attention recently, which not only inherits the advantages of NMF, but also preserves the multiway structure of tensor data. Hazan \emph{et al.} \cite{hazan2005sparse} argued that preserving the multiway structure of the image objects in the decomposition algorithm contribute to learning the localized parts of the image tensor data. Therefore, they proposed the nonnegative tensor factorization (NTF) to approximate the gray-scale image database by a linear sum of outer products of nonnegative vectors, and thus generated the part-based basis from the gray-scale image data. However, NTF assumed that the low-rank properties of each dimension of tensor data are the same \cite{he2020hyperspectral}, which go against explain the composition of the tensor data. Kim and Choi \cite{kim2007nonnegative} developed nonnegative Tucker decomposition (NTD) to introduce a core tensor which is represent the connection and interaction between different factor matrices to discover the most significant links between components and enhances the interpretability of the model, and it has been applied to image denoising \cite{kim2007nonnegative} and blind source separation \cite{zhou2012fast}. However, the existence of the core tensor in NTD increases the complexity of model calculation and estimation \cite{zhou2012fast}, which is not conducive to represent high-order tensor data. Recently, nonnegative tensor train (NTT) decomposition \cite{lee2016nonnegative} has been studied in high-order tensor analysis that is approximately represents a $d$th-order tensor as multi-linear products of $d$ low-order core tensors. The border core tensor of NTT are matrices, which is the key to reconstruct the original tensor on NTT. However, the border core tensors can only maintain relatively limited the connection and interaction with each other, which may not conducive to explain the composition of the tensor data.

The above-mentioned tensor models are similar in that both discover the multiway structure of tensor data in the high-dimensional space. However, many researchers \cite{chen2014similarity,cai2010graph} have recently shown that the high-dimensional tensor data space is generally a nonlinear manifold embedded in the low-dimensional space, and thus preserving the manifold structure of the high-dimensional space is essential for data representation. Wang \emph{et al.} \cite{wang2011image} developed a graph Laplacian regularized nonnegative tensor factorization algorithm (LRNTF) for image representation, with aims to incorporate graph regularization into NTF and take into account the manifold structure of the image space. However, LRNTF also assumes that the low-rank properties of each dimension of tensor data are the same \cite{he2020hyperspectral}, which may not conducive to tensor data representation. Jiang \emph{et al.} \cite{jiang2018image} proposed a graph-Laplacian Tucker tensor decomposition (GLTD) that considering the manifold structure of image tensor data on Tucker model for image representation and dimension reduction. GLTD does not preserved the components as nonnegative, and thus cannot be sufficiently learn localized parts of nonnegative tensor objects. Qiu \emph{et al.} \cite{qiu2019graph} proposed a graph regularization nonnegative Tucker decomposition (GNTD) algorithm for image representation, which has been shown to generate more discriminative representation of tensor data than LRNTF. However, GNTD based on Tucker structure will be exposed to the curse of dimensionality \cite{zhou2015efficient}, and thus limits it to processing the high-order tensor data.

In this paper, we propose a novel nonnegative tensor ring (NTR) decomposition to learn the localized parts of nonnegative tensor objects. To further capture the manifold geometric information of tensor data, the graph regularization nonnegative tensor ring (GNTR) decomposition is developed by combining graph regularized term with NTR. Specifically, the main contributions of this paper can be summarized as follows:
\begin{table}[t]
	\caption{List of the notation}
	\label{notation}       
	\scalebox{1}{
		\begin{tabular}{ll|ll}
			\hline
			Notations & Descriptions & Notations & Descriptions  \\
			\hline 
			$\mathbf{X}$ & A matrix & $\mathcal{X}$ & A tensor \\ 
			$\mathbf{I}$ & Identity matrix &  $\operatorname{Tr}\{\cdot\}$ & Trace\\
			$\textbf{x}$ & a vector & $\|\cdot\|_{F}$ & Frobenius norm\\
			$\circ$ & Outer product & $*$ & Hadamard product\\
			$\odot$ & Khatri–Rao product &$\otimes$ & Kronecker product\\
			\hline
	\end{tabular}}
\end{table}

\begin{itemize}
	\item [(1)] NTR expresses the information of each dimension of tensor data by the corresponding 3rd-order core tensors to learn the localized parts of nonnegative tensor objects and provide more interpretable and meaningful representation.
	\item [(2)] To further capture the manifold geometric information of tensor data with NTR, we develop GNTR that explicitly taken into account the manifold structure of tensor data, which is modeled by incorporating the similarity information of data. Therefore, GNTR is able to achieve better performance in clustering and classification tasks of tensor data.
	\item [(3)] We develop an efficient iterative algorithm based on the accelerated proximal gradient method to efficiently optimize the NTR and GNTR models and proved its convergence property theoretically.
	\item [(4)] The experimental results demonstrated that the NTR and GNTR algorithms can extract the parts-based basis with rich colors and rich lines from the tensor objects to provide a physically meaningful and more interpretable representation, and the GNTR algorithm achieves better performance than the state-of-the-art algorithms in the clustering and classification tasks.
\end{itemize}

The rest of this paper is organized as follows: In Section \ref{sec:2} the notation and preliminaries are introduced. In Section \ref{sec:3}, NTR and GNTR algorithms are developed. Finally, the simulations on five public databases are presented in Section \ref{sec:4} and followed by conclusion in Section \ref{sec:5}.

\section{Notations and preliminaries}
\label{sec:2}
\subsection{Notation}
\label{sec:2.1}

\begin{figure}[t]
	\centering  
	\subfigure[ ]{
		\includegraphics[width=0.30\textwidth]{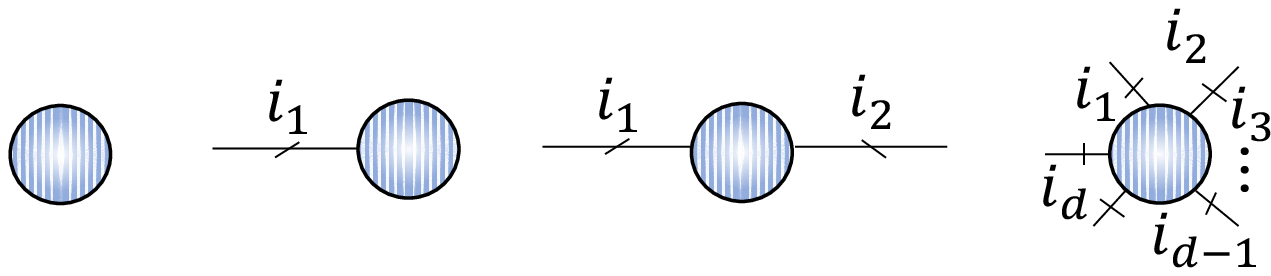}}
	\subfigure[ ]{
		\includegraphics[width=0.25\textwidth]{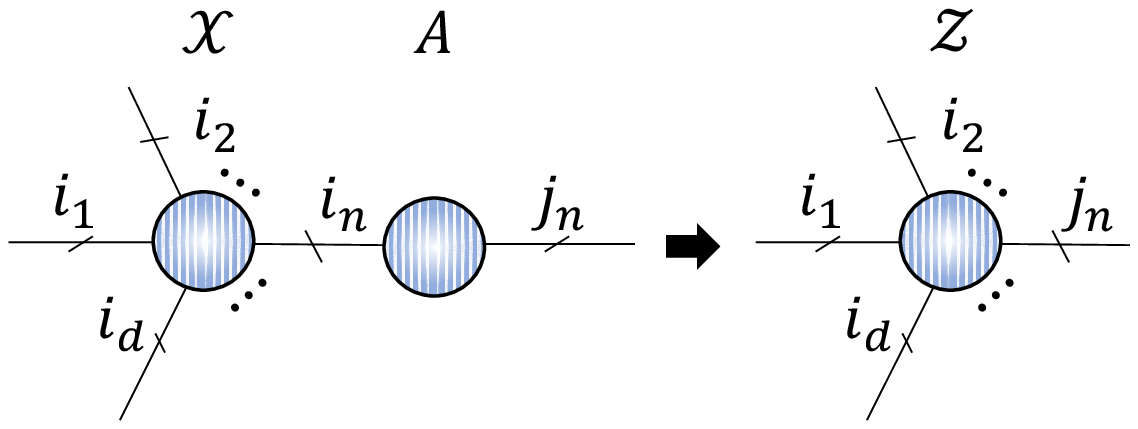}}
	\subfigure[ ]{
		\includegraphics[width=0.15\textwidth]{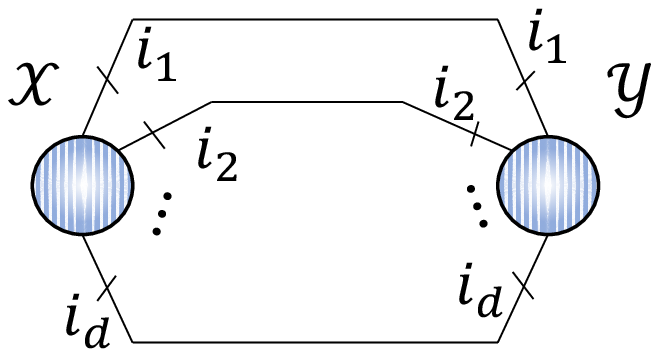}}
	\subfigure[ ]{
		\includegraphics[width=0.30\textwidth]{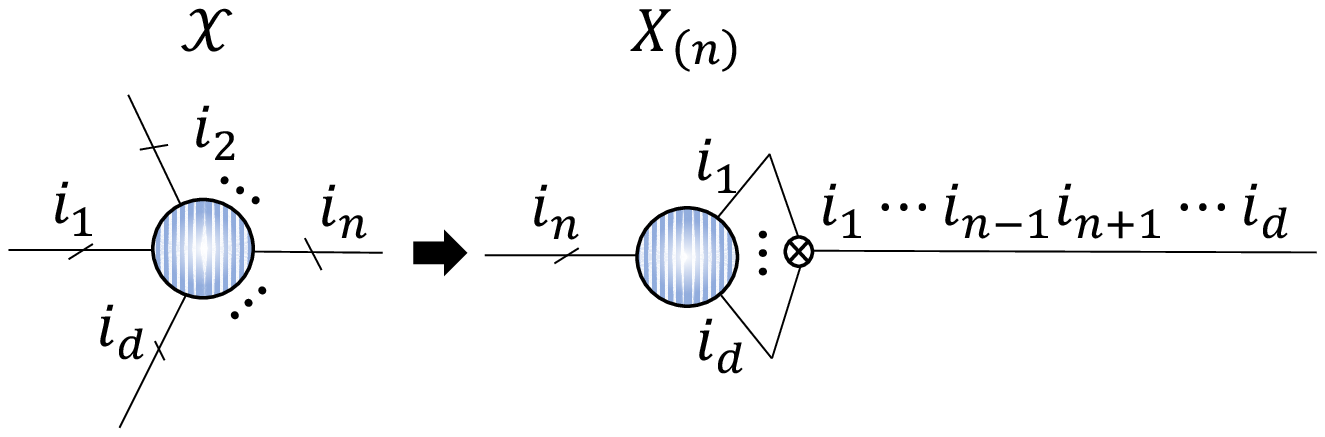}}
	\subfigure[ ]{
		\includegraphics[width=0.16\textwidth]{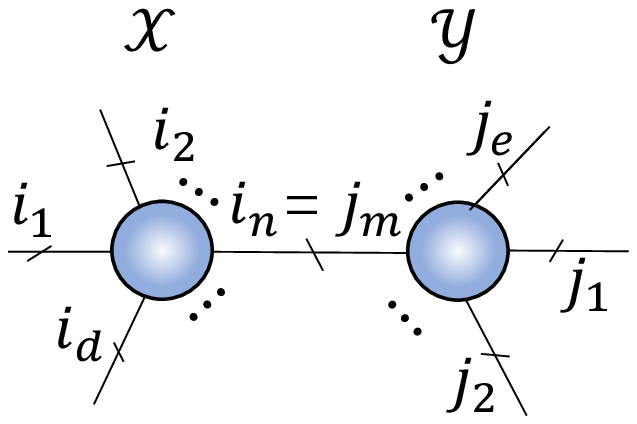}}
	\caption{Graphical representation of basic symbols and tensor operations by tensor network diagram. (a) Scalar, vector, matrix, and $d$th-order tensor. (b) Inner product. (c) Mode-$n$ product. (d) The classical mode-$n$ unfolding. (e) Contracted product}
	\label{fig:1}
\end{figure}

We review the related definitions of nonnegative tensor decomposition as follows, and the basic notations are briefly reviewed in Table.\ref{notation}.

\textbf{Definition 1} (Inner product) Given tensors $\mathcal{X}$ and $\mathcal{Y}$ of the same size $i_{1} \times i_{2} \times \cdots \times i_{d}$, the inner product of $\mathcal{X}$ and $\mathcal{Y}$ is the sum of the products of their entries, and can be written as
\begin{equation}
	\label{eq1}
	z = \left \langle \mathcal{X},\mathcal{Y} \right \rangle = \sum_{m_{1}, \ldots, m_{d}=1}^{i_{1}, \ldots, i_{d}} {\mathcal{X}}_{m_{1}, \ldots, m_{d}} {\mathcal{Y}}_{m_{1}, \ldots, m_{d}}.
\end{equation}

\textbf{Definition 2} (Mode-$n$ product) The mode-$n$ product of a tensor $\mathcal{X}$ with a matrix $\mathbf{A} \in \mathcal{R}^{j_{n} \times i_{n}}$ is denoted by $\mathcal{Z}=\mathcal{X} \times_n \mathbf{A} \in \mathcal{R}^{i_{1} \times \cdots \times i_{n-1} \times j_{n} \times i_{n+1} \times \cdots \times i_{d}}$. Elementwise, we have
\begin{equation}
	\label{eq2}
	\mathcal{Z}_{i_{1}\ldots i_{n-1}j_{n}i_{n+1}\ldots i_{d}}= \sum_{m_{n}=1}^{i_{n}} x_{i_{1}i_{2}\ldots i_{n-1}m_{n}i_{n+1}\ldots i_{d}}a_{j_{n} m_{n}}.
\end{equation}

\textbf{Definition 3} (Mode-$n$ unfolding) The mode-$n$ unfolding of a tensor $\mathcal{X}$ is a matrix defined by fixing all the indices except $i_{n}$ corresponds to a specific dimensional permutation on $\mathcal{X}$, which denotes $\mathbf{X}_{[n]} \in \mathcal{R}^{i_{n} \times i_{n+1} \cdots i_{d} i_{1} \cdots i_{n-1}}$. The classical mode-$n$ unfolding of a tensor $\mathcal{X}$ is a matrix obtained by fixing all the indices except $i_{n}$, which is denoted by $\mathbf{X}_{(n)} \in \mathcal{R}^{i_{n} \times i_{1} \cdots i_{n-1} i_{n+1} \cdots i_{d}}$. See \cite{kolda2009tensor} \cite{zhao2016tensor} for details.

\textbf{Definition 4} (Contracted product)
Given tensors $\mathcal{X} \in \mathcal{R}^{i_{1} \times i_{2} \times \dots \times i_{n} \times \dots \times i_{d}}$ and $\mathcal{Y} \in \mathcal{R}^{j_{1} \times j_{2} \times \dots \times j_{m} \times \dots \times j_{e}}$, and $i_{n}=j_{m}$, the contracted product of two tensors is denoted by a tensor $\mathcal{Z} \in \mathcal{R}^{i_{1} \times \dots \times i_{n-1} \times i_{n+1} \times \dots \times i_{d} \times j_{1} \times \dots \times j_{m-1} \times j_{m+1} \times \dots \times j_{d}}$. Elementwise, we have
\begin{equation}
	\label{eq3}
	\begin{aligned}
		&\mathcal{Z}_{i_{1}, \dots, i_{n-1}, i_{n+1}, \dots, i_{d}, j_{1}, \dots, j_{m-1}, j_{m+1}, \dots, j_{d}} \\
		&= \sum_{k=1}^{i_{n}} {x}_{i_{1}, \dots, i_{n-1}, k, i_{n+1}, \dots, i_{d}} {y}_{j_{1}, \dots, j_{m-1}, k, j_{m+1}, \dots, j_{d}}, 
	\end{aligned}
\end{equation}
which can be expressed as $\mathcal{T} = \mathcal{X} {\times}^{n}_{m} \mathcal{Y}$. This operation is also known as the contracted product of two tensors in a single common mode. For the special case of $n=d$ and $m=1$, for convenience, it can be rewritten as $\mathcal{T} = \mathcal{X} {\times}^{d}_{1} \mathcal{Y} = \mathcal{X} \bullet \mathcal{Y}$.

To facilitate the comparison, Fig. \ref{fig:1}. depicts the graphical representation of basic symbols and tensor operations via tensor network diagram \cite{cichocki2016tensor} \cite{cichocki2017tensor}.

\subsection{Nonnegative CP Decomposition}
\label{sec:2.2}
For the nonnegative multway representation learning, T.Hazan \emph{et al.} \cite{hazan2005sparse} proposed an NTF algorithm and proved its effectiveness in image representation. Especially, the experimental result has verified that NTF can learn the localized parts of grayscale image objects. The NTT algorithm expressed a nonnegative tensor $\mathcal{X}$ as the sum of a finite number of rank-$1$ nonnegative tensors
\begin{equation}
	\label{eq4}
	\mathcal{X}=\sum_{r=1}^{R} \textbf{a}_{r}^{(1)} \circ \textbf{a}_{r}^{(2)} \circ \cdots \circ \textbf{a}_{r}^{(d)}, 
\end{equation}
where $\mathcal{X} \in \mathbb{R}^{{i_{1}}\times{i_{2}}\times \cdots\times {i_{d}}}$ denotes the $d$-order tensor data. The vector $\textbf{a}_{r}^{(n)} \in \mathbb{R}^{i_{n}},(n=1,2, \cdots,d)$ and $\textbf{a}_{r}^{(n)} \geq 0$. The scalar $r$ is the nonnegative rank of NTF.

As presented in \cite{he2020hyperspectral}, the low-rank property of data is treated equally with a much larger nonnegative rank $r$ in NTF. However, the low-rank property of each dimension of real data is usually inconsistent, and thus NTF not be able to effectively explain the composition of these data.

\begin{figure}[!t]
	\centering  
	\subfigure[ ]{
		\includegraphics[width=0.20\textwidth]{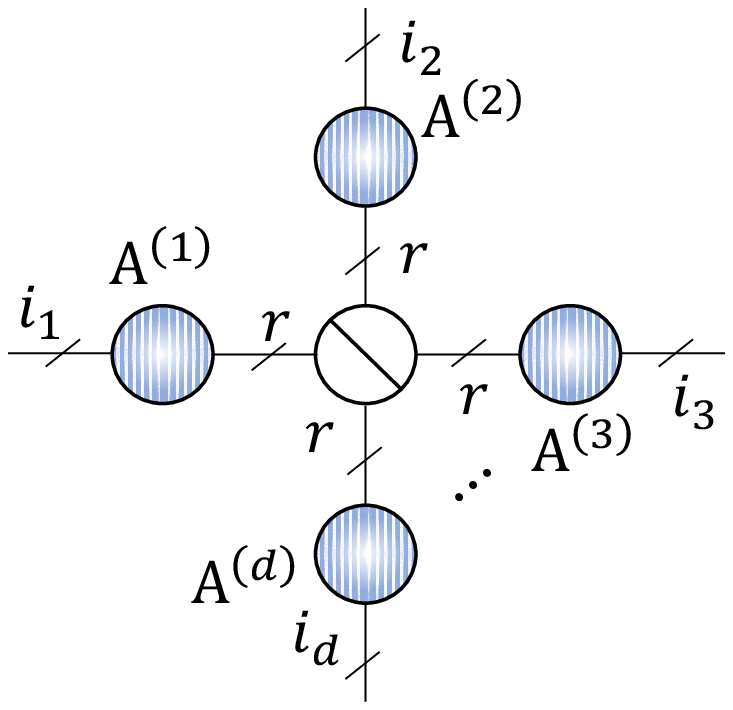}}
	\subfigure[ ]{
		\includegraphics[width=0.20\textwidth]{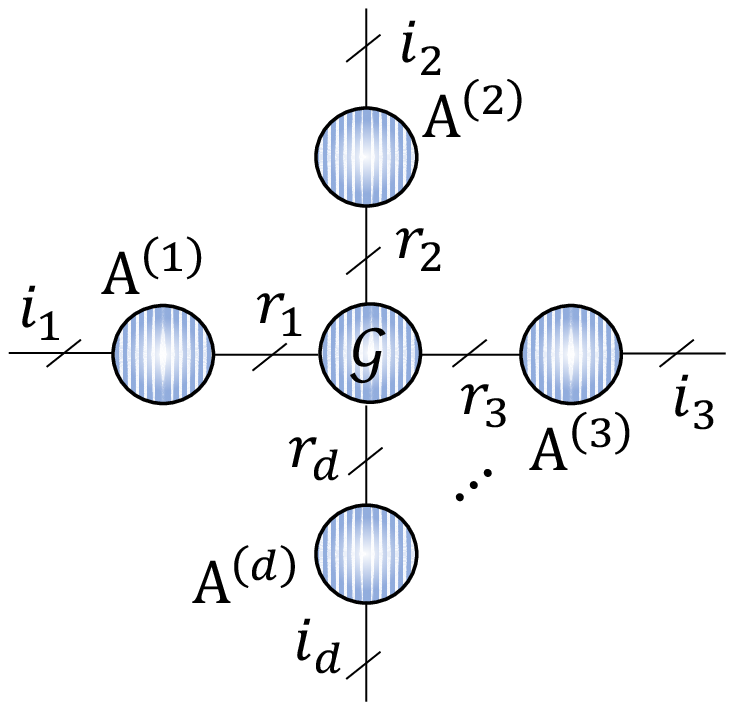}}
	\subfigure[ ]{
		\includegraphics[width=0.20\textwidth]{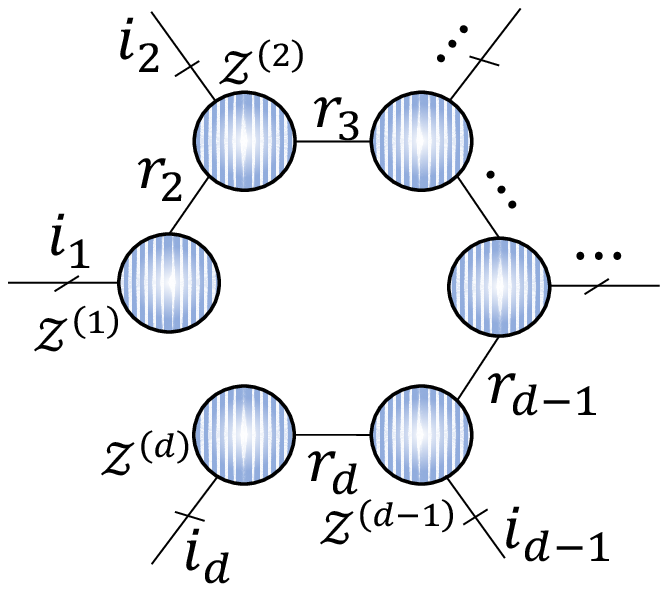}}
	\subfigure[ ]{
		\includegraphics[width=0.20\textwidth]{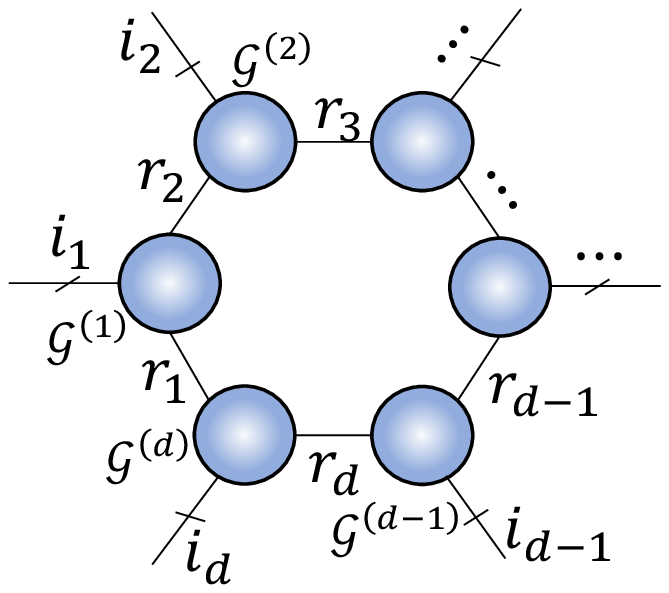}}
	\caption{Graphical representation of tensor decomposition models by tensor network diagram. (a) The CP decomposition. (b) The Tucker decomposition. (c) The Tensor Train decomposition. (d) The Tensor Ring decomposition.}
	\label{fig:2}
\end{figure}

\subsection{Nonnegative Tucker Decomposition}
\label{sec:2.3}
NTD \cite{kim2007nonnegative} is another popular nonnegative tensor decomposition for nonnegative multiway representation learning. Given a nonnegative tensor $\mathcal{X}$, NTD represents it as a nonnegative core tensor multiplied by nonnegative factor matrices in each mode, which can be achieved by solving the following problem
\begin{equation}
	\label{eq8}
	\begin{aligned}
		\mathcal{X}=\mathcal{G} \times_{1} \mathbf{A}^{(1)} \times_{2} \mathbf{A}^{(2)} \cdots \times_{n} \mathbf{A}^{(d)}
	\end{aligned}
\end{equation}
where $\mathcal{G} \in \mathbb{R}^{r_{1} \times r_{2} \times \cdots \times r_{d}}$ denotes the nonnegative core tensor and $R = \left[r_{1}, r_{2}, \cdots, r_{d}\right]$ is the nonnegative multiway rank of NTD. $A_{(n)} \in \mathbb{R}^{i_{n} \times r_{n}},(n=1,2, \cdots, d)$ denotes the nonnegative factor matrices.

NTD discovers the most significant links between components by connecting each factor matrix with a core tensor and thus provides interpretability of the model. However, the existence of the core tensor also increases the complexity of the model for computation and estimation \cite{he2020hyperspectral} that makes NTD have the limitations for high-order tensors representation.

\subsection{Nonnegative Tensor Train Decomposition}
\label{sec:2.4}
Tensor train (TT) decomposition is developed by Oseledets \emph{et al.} \cite{oseledets2011tensor} in the numerical analysis community, which is widely used in image completion \cite{yuan2017completion} and model compression of deep neural networks \cite{bengua2017matrix}. Lee \emph{et al.} extended TT to the nonnegative TT (NTT) \cite{lee2016nonnegative} and verified NTT-Tucker (NTT combined with NTD) achieves high clustering performance with lower storage cost than NTD for image representation. Nonnegative tensor $\mathcal{X}$ is expressed as the contracted product of nonnegative low-order core tensors by NTT
\begin{equation}
	\label{eq11}
	\mathcal{X}=\mathcal{Z}^{(1)} \bullet \mathcal{Z}^{(2)} \bullet \cdots \bullet \mathcal{Z}^{(d)}, 
\end{equation}
where $\mathcal{Z}^{(n)} \in \mathbb{R}^{r_{n} \times i_{n} \times r_{n+1}},(n=1,2, \cdots, d)$ is denoted by the $n$-th nonnegative core tensors. $R = \left[r_{1}, r_{2}, \cdots, r_{d}\right]$ is the nonnegative multiway rank of NTT and it is worth mentioning that $r_{1} = r_{d+1} = 1$ is a constraint of NTT that the border core tensors of NTT are both matrices as shown in Fig. \ref{fig:2}. 

NTT based on the TT structure is suitable for high-order tensor data representation due to its efficient data compression. Unfortunately, the border core tensors can only maintain relatively limited the connection and interaction with each other, and thus it may not conducive to interpreting the composition of the tensor data.

\section{Graph Regularized Nonnegative Tensor Ring Decomposition}
\label{sec:3}
In this section, we first propose NTR model based on TR structure \cite{zhao2016tensor} that aims to represent the data as the circular contractions of nonnegative core tensors. Secondly, we developed the GNTR model, which inherits the advantages of NTR while capturing the manifold geometric information. Finally, we developed a fast and efficient iterative algorithm based on the accelerated proximal gradient method to optimize the NTR and GNTR model.

\subsection{Tensor Ring Decomposition}
Tensor ring (TR) \cite{zhao2019learning} is a more general decomposition model than TT, which has been recently shown to be more powerful and efficient in various applications, e.g., tensor-based image completion \cite{yuan2019tensor}, hyperspectral image denoising \cite{chen2019nonlocal}, deep multi-modal feature fusion \cite{hou2019deep}, neural network compression \cite{wang2018wide}, etc. A high-order tensor can be represented as the circular contractions over a sequence of 3rd-order core tensor as shown in Fig. \ref{fig:2}. Each element of the tensor $\mathcal{X}$ can be written the following TR format
\begin{equation}
	\label{eq15}
	\mathcal{T}\left(i_{1}, i_{2}, \ldots, i_{d}\right)=\operatorname{Tr}\left\{\prod_{n=1}^{d} \mathbf{G}_{n}(i_{n})\right\}, 
\end{equation}
where $\mathbf{G}_{n}(i_{n}) \in \mathbf{R}^{r_{n} \times r_{n+1}}$ is the $i_{n}$-th lateral slice matrix of the core tensor $\mathcal{G}_{n} \in \mathbf{R}^{r_{n} \times i_{n} \times r_{n+1}}$. $\mathbb{R} = \left[r_{1}, r_{2}, \cdots, r_{d}\right], r_{1}=r_{d+1}$ denotes TR multiway rank and the rank constraint. TR restricts one dimension of the border core tensors to be equal, that is, both border core tensors are 3$rd$-order tensor. The border core tensors can maintain the direct connection and interactions with each other due to the circular dimensional permutation invariance of the TR model \cite{zhao2016tensor}. The graphical representation Fig. \ref{fig:2}. can show this structure more clearly.

\subsection{Objection Function of NTR}
Previous works \cite{yuan2019tensor,chen2019nonlocal,hou2019deep,wang2018wide} have demonstrated that TR is a powerful tool for processing high-order tensor data. However, they cannot sufficiently learn localized parts of tensor objects to further interpret the composition of the tensor data. To explain the composition of the data more clearly, NTR based on the TR structure is developed, which is defined by solving the following problem:
\begin{equation}
	\begin{aligned}
		\label{eq16}
		\min_{\mathcal{G}^{(n)}}&\left\|\mathcal{X}-\operatorname{NTR}\left(\mathcal{G}^{(1)}, \mathcal{G}^{(2)}, \cdots, \mathcal{G}^{(d)}\right)\right\|^{2}_{F} \\
		&\text{s.t.}\ \mathcal{G}^{(n)} \geq 0,(n=1,2, \cdots, d), 
	\end{aligned}
\end{equation}
where $\mathcal{G}^{(n)} \in \mathbb{R}^{r_{n} \times i_{n} \times r_{n+1}}$ denotes $n$-th nonnegative core tensor. $\operatorname{NTR}\left(\mathcal{G}^{(1)}, \mathcal{G}^{(2)}, \cdots, \mathcal{G}^{(d)}\right)$ is defined as the reconstruction of the original tensor through the circular contraction of the core tensors, and can be rewritten as the following mode-$n$ matrix form
\begin{equation}
	\begin{aligned}
		\label{eq17}
		\mathbf{X}_{[n]}=\mathbf{G}_{(2)}^{(n)}\left(\mathbf{G}_{\left[ 2 \right]}^{\neq n}\right)^{\top}, 
	\end{aligned}
\end{equation}
where $\mathbf{G}_{(2)}^{(n)} \in \mathbb{R}^{i_{n} \times r_{n} r_{n+1}}$ denotes the mode-$2$ unfolding core tensor. $\mathbf{G}^{\neq n}_{\left[ 2 \right]} \in \mathbb{R}^{i_{n+1} \cdots i_{d}i_{1} \cdots i_{n} \times r_{n}r_{n+1}}$ is defined as a mode-2 unfolding matrix of subchain tensor $\mathcal{G}^{\neq n} \in \mathbb{R}^{r_{n+1} \times i_{n+1} \cdots i_{d}i_{1} \cdots i_{n} \times r_{n}}$. The subchain tensor $\mathcal{G}^{\neq n}$ is obtained by merging all core tensors except the $n$-th core tensor. The problem (\ref{eq16}) can be rewritten as:
\begin{equation}
	\begin{aligned}
		\label{eq18}
		&\mathcal{F}_{NTR}^{(n)}= \frac{1}{2}\left\|\mathbf{X}_{[n]}-\mathbf{G}_{(2)}^{(n)}\left(\mathbf{G}_{\left[ 2 \right]}^{\neq n}\right)^{\top}\right\|_{F}^{2} 
		\\ 
		&\text {s.t. } \mathbf{G}^{(n)} \geq 0, \mathbf{G}_{\left[ 2 \right]}^{\neq n} \geq 0,(n=1,2, \cdots, d), 
	\end{aligned}
\end{equation}
where $R = \left[r_{1}, r_{2}, \cdots, r_{d}\right]$ denotes nonnegative multiway rank of NTR and $r_{1}=r_{d+1}$ is the key to NTR successfully recovering the orginal tensor, which is essentially different from the nonnegative multiway rank constraint $r_{1}=r_{d+1}=1$ of NTT.

\subsection{Optimization of NTR}
\label{sec:3.3}
In recent years, Accelerated Proximate Gradient (APG) method has been applied to efficiently solve NMF problem, which depends on Lipschitz continuity of the gradients. We derive the respective gradients of $\mathcal{F}_{N T R}^{(n)}$ with respect to $\mathbf{G}^{(n)}_{(2)}$ as follow
\begin{equation}
	\label{eq19}
	\frac{\partial \mathcal{F}_{N T R}^{(n)}}{\partial \mathbf{G}^{(n)}_{(2)}}=\mathbf{G}^{(n)}_{(2)}\left(\mathbf{G}_{\left[ 2 \right]}^{\neq n}\right)^{\top} \mathbf{G}_{\left[ 2 \right]}^{\neq n}-\mathbf{X}_{[n]} \mathbf{G}_{\left[ 2 \right]}^{\neq n}. 
\end{equation}

It is straightforward to verify that the following two propositions hold:

\textbf{Proposition 1:} The objection function of each subproblem $\mathcal{F}_{NTR}^{(n)}$ is convex.

\textbf{Proposition 2:} The gradient (\ref{eq19}) is Lipschitz continuous with the Lipschitz constant $L_{NTR}=\left\| \left(\mathbf{G}^{\neq n}_{\left[ 2 \right]}\right)^{\top} \mathbf{G}^{\neq n}_{\left[ 2 \right]} \right\|_{2}$.

The proofs of Proposition $1$ and $2$ are presented in \cite{qiu2020generalized}. The objection function (\ref{eq18}) is non-convex, and can not obtain a global optimal solution. Fortunately, for each separable variable, the objective function of each subproblem $\mathcal{F}_{NTR}^{(n)}$ is convex according to Proposition $1$. Hence, we construct two sequences and alternatively update separable variable in each iteration round of one subproblem. Firstly, we can define the proximal function of $\mathcal{F}_{N T R}^{(n)}$ at $\mathbf{G}^{(n)}_{(2)}$ as follows:

\begin{equation}
	\begin{aligned}
		\label{eq20}
		\phi\left(\mathbf{G}_{(2)}^{(n)}, Y^{t}\right)&=\mathcal{F}_{N T R}^{(n)}+\left\langle\frac{\partial \mathcal{F}_{N T R}^{(n)}}{\partial \mathbf{G}_{(2)}^{(n)^{t-1}}}, \mathbf{G}_{(2)}^{(n)}-\mathbf{Y}^{t}\right\rangle\\
		&+\frac{L_{NTR}}{2}\left\|\mathbf{G}_{(2)}^{(n)}-\mathbf{Y}^{t}\right\|_{F}^{2}, 
	\end{aligned}
\end{equation}
where $t$ denotes the inner iteration numbers and $\left\langle \cdot, \cdot \right\rangle$ is the inner product. At the same time, the following sequence is constructed to select the search point $\mathbf{Y}^{t+1}$ as follows
\begin{equation}
	\label{eq21}
	\mathbf{Y}^{t+1}={\mathbf{G}^{(n)}_{(2)}}^{t}+\frac{\alpha_{t}-1}{\alpha_{t+1}}\left({\mathbf{G}^{(n)}_{(2)}}^{t}-{\mathbf{G}^{(n)}_{(2)}}^{t-1}\right), 
\end{equation}
where $\mathbf{Y}^{t+1}$ is denoted as the search point constructed by linearly combining the two latest approximate solutions ${\mathbf{G}^{(n)}_{(2)}}^{t}$ and ${\mathbf{G}^{(n)}_{(2)}}^{t-1}$. The combination coefficient $\alpha_{t}$ is given as

\begin{equation}
	\label{eq22}
	\alpha_{t+1}=\frac{1+\sqrt{4\alpha_{t}^{2}+1}}{2}.
\end{equation}

The approximate solution of $\mathbf{G}_{(2)}^{(n)^{t}}$ in the iteration $t$ can be obtained by minimizing the proximal function (\ref{eq20}) as follows

\begin{equation}
	\begin{aligned}
		\label{eq23}
		\mathbf{G}_{(2)}^{(n)^{t+1}}= \underset{\mathbf{G}_{(2)}^{(n)}\geq 0}{\arg\min}\phi\left(\mathbf{G}_{(2)}^{(n)^{t}}, Y\right).
	\end{aligned}
\end{equation}

By using the Lagrange Multiplier Method, the Karush-Kuhn-Tucker conditions of problem (\ref{eq23}) can be expressed in the following form

\begin{equation}
	\begin{aligned}
		\label{eq24}
		\frac{\partial \phi\left(\mathbf{G}_{(2)}^{(n)^{t}}, Y^{t}\right)}{\partial \mathbf{G}_{(2)}^{(n)}} &\geq 0, \\
		\mathbf{G}_{(2)}^{(n)^{t}} &\geq 0, \\
		\frac{\partial \phi\left(\mathbf{G}_{(2)}^{(n)^{t}}, Y^{t}\right)}{\partial \mathbf{G}_{(2)}^{(n)}} * \mathbf{G}_{(2)}^{(n)^{t}} &= 0, 
	\end{aligned}
\end{equation}
where $*$ denotes Hadamard product. By solving the problem (\ref{eq24}), the update formula for $\mathbf{G}_{(2)}^{(n)^{t}}$ is given as

\begin{equation}
	\begin{aligned}
		\label{eq25}
		\mathbf{G}_{(2)}^{(n)^{t+1}} \leftarrow \mathcal{P}_{+}\left( Y^{t}-\frac{1}{L_{NTR}}\frac{\partial \mathcal{F}_{NTR}^{(n)}}
		{\partial \mathbf{G}_{(2)}^{(n)}}\right), 
	\end{aligned}
\end{equation}
where $\mathcal{P}_{+}\left( \mathbf{X} \right)$ denotes projects the negative elements of $\mathbf{X}$ to zero. Updating $\mathbf{G}_{(2)}^{(n)}$ iteratively by Eq.(\ref{eq21}), Eq.(\ref{eq22}) and Eq.(\ref{eq25}) until convergence criterion is reached, and applying similar procedures to all the core tensors, we can obtain the solution of NTR problem (\ref{eq16}) based on APG method as shown in Algorithm \ref{NTR_algorithm}.

\begin{algorithm}[tbp] 
	\caption{NTR based on APG method}
	\label{NTR_algorithm}
	\begin{algorithmic}[1]
		\REQUIRE Tensor $\mathcal{X} \in \mathbb{R}^{i_{1} \times i_{2} \times \cdots \times i_{d}}$, nonnegative multiway rank $r_{n}$ for $ n=1, \cdots, d$, maximum number of iterations of $t_{max}$.\\ 
		\ENSURE Core tensors $\mathcal{G}_{n} $ for $ n=1,\cdots,d$.\\
		\STATE{Initialize $ \mathcal{G}_{n} \in \mathbb{R}^{r_{n} \times i_{n} \times r_{n+1}}$ for $n=1, \cdots, d$ as random tensors from Gaussian distribution}.
		\REPEAT 
		\FOR{$n$ = 1 to $d$}
		\STATE Obtain mode-2 unfolding matrix of subchain $\mathbf{G}_{[2]}^{ \neq n}$.
		\STATE Initialize $\alpha=1$, $\mathbf{Y}^{0}=\mathbf{G}_{[2]}^{ \neq n}$, $L_{NTR}=\left\|{(\mathbf{G}_{[2]}^{ \neq n})}^{\top} {\mathbf{G}_{[2]}^{ \neq n}}\right\|_{2}$.
		\FOR{$t$ = 0 to $t_{max}-1$}
		\STATE Updating ${\mathbf{G}_{(2)}^{(n)}}^{t+1}$, $\mathbf{Y}^{(t+1)}$ and ${\alpha}_{t+1}$ by using  Eq.(\ref{eq25}), Eq.(\ref{eq21}) and Eq.(\ref{eq22}) respectively.
		\ENDFOR
		\STATE {Tensorization of mode-2 unfolding matrix}\\$\mathcal{G}_{n} \leftarrow$ folding$\left({\mathbf{G}_{(2)}^{(n)}}^{t_{max}}\right)$.
		\ENDFOR
		\UNTIL{convergence}.
	\end{algorithmic}
\end{algorithm}

\subsection{Graph Regularization}
The viewpoint of manifold learning \cite{qiu2020generalized,li2016graph} take into account the observed high-dimensional data is actually mapped to a high-dimensional space by a low-dimensional manifold geometrical structure. As the limitation of the internal structure of high-dimensional data, some high-dimensional data usually produce certain redundancy of dimensions, and thus difficult to observe the manifold geometrical structure of high-dimensional data. Fortunately, the neighbor graph has been verified to effectively characterize the manifold geometrical structure \cite{belkin2002laplacian}. Based on this idea, the geometrical information is encoded by connecting each tensor subject with its $p$-nearest neighbors. The relationship matrix $\mathbf{W} \in \mathbb{R}^{n_{1} \times n_{1}}$ encodes each tensor object connections in the graph

\begin{equation}
	\label{eq26}
	\mathbf{W}_{i j}=\left\{\begin{array}{l}
		1, \text { if } \mathcal{X}_{i} \in \mathcal{N}_{p}\left(\mathcal{X}_{j}\right), \text { and } \mathcal{X}_{j} \in \mathcal{N}_{p}\left(\mathcal{X}_{i}\right) \\
		0, \text {otherwise}, 
	\end{array}\right.
\end{equation}
where $\mathcal{N}_{p}\left(\mathcal{X}_{i}\right)$ represents the set of $p$ objects closest to the tensor $\mathcal{X}_{i}$ in the graph. Many techniques \cite{cai2010graph} \cite{jiang2018image} can measure the distance between tensors, and the Frobenius norm distance is considered to simplify the problem in this case.

\subsection{Objection Function of GNTR}
To enable NTR to observe the manifold geometrical structure of high-dimensional data, we propose a graph regularized nonnegative tensor ring decomposition (GNTR). GNTR not only inherits the advantages of NTR, but also additionally learns the manifold geometric information of tensor data to enhance the effectiveness in data representation in clustering and classification tasks. GNTR incorporates the manifold geometric information of high-dimension space in NTR by minimizing the following objective function:

\begin{equation}
	\begin{aligned}
		\label{eq27}
		\mathcal{F}_{GNTR}^{(n)}=& \frac{1}{2}\left\|\mathbf{X}_{[n]}-\mathbf{G}_{(2)}^{(n)}\left(\mathbf{G}_{\left[ 2 \right]}^{\neq n}\right)^{\top}\right\|_{F}^{2}\\
		&+ \frac{\beta}{2} \operatorname{tr}\left(\left(\mathbf{G}_{(2)}^{(d)}\right)^{\top} \mathbf{H}_{g} \mathbf{G}_{(2)}^{(d)}\right)
		\\ 
		\text {s.t. } \mathbf{G}^{(n)}_{(2)} \geq 0, &\mathbf{G}_{(2)}^{\neq n} \geq 0, \mathbf{H}_{g} \geq 0, (n=1,2, \cdots, d), 
	\end{aligned}
\end{equation}
where the $\beta \geq 0$ is the parameter to control the intensity of graph regularization term. As the $d$-th dimension of $\mathcal{X}$ has defaulted the number of tensor objects, the similar information of the tensor objects is integrated into the mode-2 unfolding matrix of the $d$-th core tensor through the Laplacian matrix $\mathbf{H}_{g}=\mathbf{D}-\mathbf{W} \in \mathbb{R}^{i_{1} \times i_{1}}$ where $\mathbf{D}_{i i}=\sum_{j} \mathbf{W}_{i j}$.

\subsection{Optimization of GNTR}
In this section, we design the APG method to solve GNTR problem. We derive the gradient of $\mathcal{F}_{GNTR}^{(n)}$ with respect to $\mathbf{G}_{(2)}^{(n)}$  in the case of $n=d$ as follow
\begin{equation}
	\label{eq28}
	\frac{\partial \mathcal{F}_{G N T R}^{(n)}}{\partial \mathbf{G}^{(n)}_{(2)}}=\mathbf{G}^{(n)}_{(2)}\left(\mathbf{G}_{\left[ 2 \right]}^{\neq n}\right)^{\top} \mathbf{G}_{\left[ 2 \right]}^{\neq n}-\mathbf{X}_{[n]} \mathbf{G}_{\left[ 2 \right]}^{\neq n} + \beta \mathbf{H}_{g} \mathbf{G}^{(n)}_{(2)}. 
\end{equation}

It is simple to verify that the following propositions are true:

\textbf{Proposition 3:} The objection function of each subproblem $\mathcal{F}_{GNTR}^{(n)}$ is convex.

\textbf{Proposition 4:} For the case of $n=1$, the gradient (\ref{eq28}) is Lipschitz continuous with the Lipschitz constant $L_{GNTR}=\left\| \left(\mathbf{G}^{\neq n}_{\left[ 2 \right]}\right)^{\top} \mathbf{G}^{\neq n}_{\left[ 2 \right]} \right\|_{2}+\left\|\beta \mathbf{H}_{g} \right\|_{2}$.

The proofs of Proposition 3 and 4 are presented in \cite{qiu2020generalized}. According to the above analysis, the GNTR algorithm based on the APG method can be developed.

The manifold regularization term only affects the separable variables of GNTR in the case of $n=d$, that is, the core tensor $\mathcal{G}^{(d)}$ learns the manifold geometry information of the tensor data. In summary, we can divide the objective function into two situations for discussion.

In the case of $n=d$, it is easy to get inspiration from Section (\ref{sec:3.3}). The proximal function of $\mathcal{F}_{G N T R}^{(d)}$ at $\mathbf{G}^{(d)}_{(2)}$ is defined by

\begin{equation}
	\begin{aligned}
		\label{eq29}
		\phi\left(\mathbf{G}_{(2)}^{(d)}, Y^{t}\right)&=\mathcal{F}_{G N T R}^{(d)}+\left\langle\frac{\partial \mathcal{F}_{G N T R}^{(d)}}{\partial \mathbf{G}_{(2)}^{(d)^{t-1}}}, \mathbf{G}_{(2)}^{(d)}-\mathbf{Y}^{t}\right\rangle\\
		&+\frac{L_{GNTR}}{2}\left\|\mathbf{G}_{(2)}^{(d)}-\mathbf{Y}^{t}\right\|_{F}^{2}, 
	\end{aligned}
\end{equation}
where the search point $\mathbf{Y}^{t}$ is obtained by Eq.(\ref{eq21}) and the combination coefficient $\alpha$ is defined by Eq.(\ref{eq22}). Then $\mathbf{G}_{(2)}^{(d)^{t}}$ is obtained by minimizing the function (\ref{eq29}) under $\mathbf{G}^{(d)}_{(2)} \geq 0$, as shown in

\begin{equation}
	\begin{aligned}
		\label{eq30}
		\mathbf{G}_{(2)}^{(d)^{t+1}} \leftarrow\mathcal{P}_{+}\left( \mathbf{Y}^{t}-\frac{1}{L_{GNTR}}\frac{\partial \mathcal{F}_{GNTR}^{(d)}}
		{\partial \mathbf{G}_{(2)}^{(d)^{t}}}\right). 
	\end{aligned}
\end{equation}

In the case of $n=d$, the $\mathcal{G}^{(d)}$ can be updated by Eq.(\ref{eq21}), Eq.(\ref{eq22}) and Eq.(\ref{eq30}) until convergence criterion is reached. In the case of $n=1, \cdots, d-1$, we can obtained the $\mathcal{G}^{(n)}$ by updating the Eq.(\ref{eq21}), Eq.(\ref{eq22}) and Eq.(\ref{eq25}) until convergence criterion is reached. In summary, the solution of GNTR problem can solved by using APG method as shown in Algorithms \ref{GNTR_algorithm}.

\subsection{Convergence Analysis}
Because problem (\ref{eq16}) is non-convex, the global optimal solution cannot be obtained. Paatero \cite{paatero1999multilinear} proved that for the alternating nonnegative least squares method, no matter how many block variables there are, the method can only converge if there is a unique solution in each sub-problem. We will prove the problem (\ref{eq16}) can guaranteed the weak convergence property under some mild conditions, which is similar to the proof of \cite{qiu2020generalized}.

\textbf{Proposition 5:} Suppose $\mathcal{X} \in \mathbb{R}^{i_{1} \times i_{2} \times \cdots \times i_{d}}$ is defined by a $d$th-order tensor with TR rank $\left[ r_{1} ,\cdots, r_{d} \right]$, $r_{1}=r_{d+1}$ and $r_{n}r_{n+1} \leq i_{n}$. Then we have a TR decomposition of $\mathcal{X}$, i.e., $\mathcal{X}=\operatorname{TR} \left(\mathcal{G}_{1}, \cdots, \mathcal{G}_{d} \right)$,  and its mode-$n$ unfolding matrix $\mathbf{X}_{[n]}, \left( n = 1, \cdots, d \right)$ is full rank. The mode-$n$ unfolding matrix $\mathbf{X}_{[n]}$ is also full rank, and can represent as $\mathbf{X}_{[n]}=\mathbf{G}_{(2)}^{(n)}\left(\mathbf{G}_{\left[ 2 \right]}^{\neq n}\right)^{\top}$. The $\mathbf{G}_{\left[ 2 \right]}^{\neq n}$ is also full rank. 

\textbf{Proof 1:} To prove the proposition 5, we show that the $\mathbf{G}^{\neq n}_{(2)}$ have full rank when the mode-$2$ unfolding matrices have full rank and $r_{n}r_{n+1} \leq i_{n}$. The $d$-order tensor $\mathcal{X}$ can be represent as the TR decomposition $\operatorname{}{TR}\left(\mathcal{G}^{(1)}, \mathcal{G}^{(2)}, \cdots, \mathcal{G}^{(d)}\right)$. If the $\mathbf{G}^{\neq n}_{(2)}$ is not full rank, the core tensors can be further decomposed as follow $\mathbf{G}_{\left[ 2 \right]}^{\neq n}=\widetilde{\mathbf{G}}_{\left[ 2 \right]}^{\neq n}\mathbf{U}^{(n)} \in \mathbf{R}^{r_{n}'r_{n+1}' \times i_{1} \dots i_{d}}$ and the transfer matrix $\mathbf{U}^{(n)}\in \mathbb{R}^{r_{n}'r_{n+1}' \times r_{n}r_{n+1}}$ can be merged into the core tensor $\widetilde{\mathbf{G}}_{(2)}^{(n)}=\mathbf{G}_{(2)}^{(n)} {\mathbf{U}^{(n)}}^{\top}\in \mathbb{R}^{i_{1} \dots i_{d} \times r_{n}'r_{n+1}'}$, where $r_{n}'r_{n+1}' < r_{n}r_{n+1}$. Then, we have a new TR decomposition $\operatorname{}{TR}\left(\widetilde{\mathcal{G}}^{(1)}, \widetilde{\mathcal{G}}^{(2)}, \cdots, \widetilde{\mathcal{G}}^{(d)}\right)$ of tensor $\mathcal{X}$. This demonstrates that $\operatorname{rank}(\mathbf{X}_{n}) \leq r_{n}'r_{n+1}' < r_{n}r_{n+1}$, which contradicts the assumption. This completes the proof.

\textbf{Theorem:} Let $\mathbf{G}^{(n)}_{(2)}, \left( n = 1, \cdots, d \right)$ both are full rank, that is, $\operatorname{rank}\left( \mathbf{G}^{(n)}_{(2)} \right) = r_{n}r_{n+1}$. each subproblem $\mathcal{F}_{NTR}^{(n)}$ can converge to the unique and optimal solution, which indicates local convergence is guaranteed.

\textbf{Proof 2:}
The Hessian matrix of $\mathcal{F}_{NTR}^{(n)}$ is derived as follows
\begin{equation}
	\frac{\partial \mathcal{F}_{N T R}^{(n)}}{\partial \mathbf{G}^{(n)}_{(2)} \partial \mathbf{G}^{(n)}_{(2)}} = \mathbf{E}_{a} \otimes \left(\mathbf{G}_{\left[ 2 \right]}^{\neq n}\right)^{\top} \mathbf{G}_{\left[ 2 \right]}^{\neq n}, 
	\label{eq35}
\end{equation}
where $\mathbf{E}_{a} \in \mathbb{R}^{r_{n-1} r_{n+1} \times r_{n-1} r_{n+1}}$ denotes identity matrix. As shown in (\ref{eq35}), the $\mathcal{F}_{N T R}^{(n)}$ is strictly convex only if $\left(\mathbf{G}_{\left[ 2 \right]}^{\neq n}\right)^{\top} \mathbf{G}_{\left[ 2 \right]}^{\neq n}$ is positive definite. According to the Proposition 5, we find that the Hessian matrix of subploblem (\ref{eq16}) has full rank. Therefore, each subploblem $\mathcal{F}_{N T R}^{(n)}$ is strictly convex. This completes the proof.

The convergence of the GNTR algorithm can also be derived from the above similar method. We find that the Hessian matrix of each subproblem (\ref{eq27}) is also positive definite. Therefore, each subproblem $\mathcal{F}_{G N T R}^{(n)}$ is strictly convex, which thus the convergence of the GNTR algorithm also can be proved.

\begin{algorithm}[tbp] 
	\caption{GNTR based on APG method} 
	\label{GNTR_algorithm} 
	\begin{algorithmic}[1]
		\REQUIRE Tensor $\mathcal{X} \in \mathbb{R}^{i_{1} \times i_{2} \times \cdots \times i_{d}}$, nonnegative multiway rank $r_{n}$ for $ n=1, \cdots, d$, maximum number of iterations of $t_{max}$, balance parameter $\beta$.\\ 
		\ENSURE Core tensors $\mathcal{G}_{n} $ for $ n=1,\cdots,d$.\\
		\STATE{Initialize $ \mathcal{G}_{n} \in \mathbb{R}^{r_{n} \times i_{n} \times r_{n+1}}$ for $n=1, \cdots, d$ as random tensors from Gaussian distribution}.
		\REPEAT 
		\FOR{$n$ = 1 to $d$}
		\STATE Obtain mode-2 unfolding matrix of subchain $\mathbf{G}_{[2]}^{ \neq n}$.
		\STATE Initialize $\alpha=1$, $\mathbf{Y}^{0}=\mathbf{G}_{[2]}^{ \neq n}$, $L_{NTR}=\left\|{(\mathbf{G}_{[2]}^{ \neq n})}^{\top} {\mathbf{G}_{[2]}^{ \neq n}}\right\|_{2}$, $L_{GNTR}=L_{NTR}+\left\|\beta \mathbf{H}_{g} \right\|_{2}$.
		\FOR{$t$ = 0 to $t_{max}-1$}
		\IF{$n=d$}
		\STATE Updating ${\mathbf{G}_{(2)}^{(n)}}^{t+1}$ by using Eq.(\ref{eq30}).
		\ELSE
		\STATE Updating ${\mathbf{G}_{(2)}^{(n)}}^{t+1}$ by using Eq.(\ref{eq25}).
		\ENDIF
		\STATE Updating $\mathbf{Y}^{t+1}$ and ${\alpha}_{t+1}$ by using Eq.(\ref{eq21}) and Eq.(\ref{eq22}) respectively.
		\ENDFOR
		\STATE {Tensorization of mode-2 unfolding matrix}\\$\mathcal{G}_{n} \leftarrow$ folding$\left({\mathbf{G}_{(2)}^{(n)}}^{t_{max}}\right)$.
		\ENDFOR
		\UNTIL{convergence}.
	\end{algorithmic}
\end{algorithm}

\section{Tasks}
\label{sec:4}
In this section, we first visualized the basis extracted by each algorithm to compare the ability of different algorithms to learn localized parts of the tensor objects. Secondly, we conduct clustering and classification tasks on five publicly available real-world databases to comprehensively compare the performance of our proposed NTR and GNTR algorithms. Thirdly, the clustering tasks across different parameters of NTR and GNTR are conducted to investigate the parameter sensitivity of the algorithm. Finally, we show that the convergence curves of NTR and GNTR on five databases.

\subsection{Databases}
\label{sec:4.1}
There are in total five databases used in our tasks, and the descriptions of these databases are given as follow
\begin{itemize}
	\item ORL Database: The ORL database consists of 400 grayscale $112\times92$ face images of 40 distinct subjects. Each individual has 10 different images under different times with varying lighting, facial expressions, and facial details. We adjusted the resolution of each image to $32 \times 27$ and construct a 3rd-order data tensor $\mathcal{T} \in \mathcal{R}^{32 \times 27 \times 400}$.
	\item FEI PART 1 Database: The FEI PART 1 face database is the subset of FEI database, which consists of 700 color images of size $480 \times 640 \times 3$ collected from 50 individuals. Each individual has 14 different images under different view and facial expressions. Each image is downsampled to 48$\times$64 pixels and we can construct a $4$th-order data tensor $\mathcal{T} \in \mathcal{R}^{48 \times 64 \times 3 \times 700}$.
	\item GT Database: The Georgia Tech Face database contains 750 color images of 50 people that each image is the resolution of $640 \times 480$ pixels. The background of the image is messy, the faces have different orientations, and also have different facial expressions, lighting conditions, and proportions. We downsampled each image to $40 \times 30 \times 3$, so we have a 4th-order data tensor $\mathcal{T} \in \mathcal{R}^{40 \times 30 \times 3 \times 750}$.
	\item COIL-100 PART 1 Database: The Columbia Object Image Library (COIL-100) is a database of 7200 color images of 100 objects. Each object has 72 images of size $128 \times 128 \times 3$ taken from different poses. We only considered the first 20 categories and resized all the images into $32 \times 32 \times 3$ to obtain the COIL-100 PART 1 database, as a 4th-order data tensor $\mathcal{T} \in \mathcal{R}^{32 \times 32 \times 3 \times 1400}$.
	\item Faces94 PART 1 Database: The Faces94 database consists of 3060 color images of 153 individuals in permanent positions with respect to the camera for a total of 20 facial expressions. We used images of the first 72 individuals and downsampled them to $50 \times 45 \times 3$ pixels. Finally, we can obtain the Faces94 PART 1 database, as a 4th-order data tensor $\mathcal{T} \in \mathcal{R}^{ 50 \times 45 \times 3 \times 1440}$.
\end{itemize}

Since the permutation of database dimensions, the extracted feature from tensor database by TT, GNTT, TR, NTR, and GNTR is the last core tensor and we call it to feature core tensor.

\subsection{Comparative Algorithms}
\label{sec:4.2}
Our proposed algorithms are related to matrix/tensor decomposition algorithm and graph Laplacian regularization, so we compare our algorithms with the following state-of-the-art algorithms:
\begin{itemize}
	\item PCA \cite{wold1987principal}: Principal component analysis (PCA) is one of the most famous unsupervised dimensionality reduction algorithms.
	\item gLPCA \cite{jiang2013graph}: Graph-Laplacian principal component analysis (gLPCA) denotes PCA algorithm by considering the manifold structure of the data.
	\item NMF \cite{lee1999learning}: Nonnegative matrix factorization aims to learn the nonnegative parts-based basis of data objects.
	\item GNMF \cite{cai2010graph}: Graph regularized nonnegative matrix factorization (GNMF) denotes NMF considering the manifold geometric structure in the data.
	\item NTF \cite{hazan2005sparse}:  Nonnegative CANDECOMP/PARAFAC decomposition denotes a nonnegative representation algorithm of tensor data based on CP decomposition.
	\item LRNTF \cite{wang2011image}: Laplacian regularized nonnegative tensor factorization is defined by NTF algorithm with considering the manifold structure of the data.
	\item GLTD \cite{jiang2018image}: Graph-Laplacian tucker tensor decomposition (GLTD) denotes the unconstrained Tucker model combined with graph regularized term.
	\item NTD \cite{kim2007nonnegative}: Nonnegative Tucker decomposition denotes a nonnegative representation algorithm based on Tucker structure.
	\item GNTD \cite{qiu2019graph}: Graph regularized nonnegative Tucker decomposition denotes NTD algorithm by considering the manifold geometric structure of data.
	\item NTT \cite{lee2016nonnegative}: Nonnegative tensor train (NTT) decomposition represents tensor data as a series of nonnegative low-rank core tensors and the APG method is adopted to optimize the NTT model.
	\item GNTT: Graph regularized nonnegative tensor train (GNTT) decomposition denotes NTT combined with graph regularized term and the APG method is adopted to optimize the GNTT model.
\end{itemize}

\begin{figure*}[!t]
	\centering  
	\subfigure[ ]{
		\includegraphics[width=0.18\textwidth]{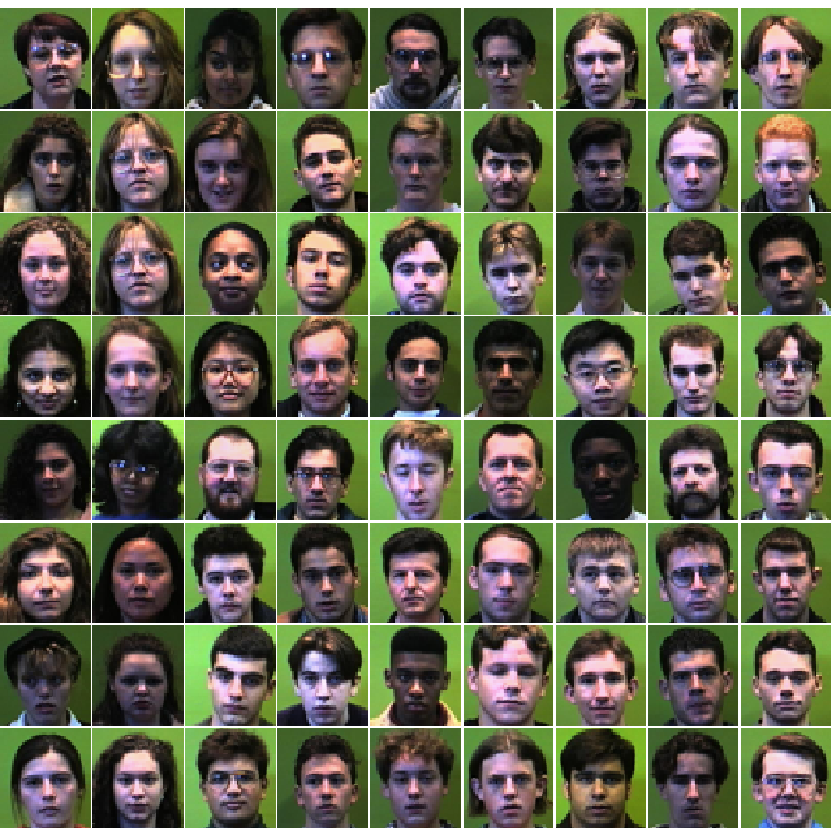}}
	\subfigure[ ]{
		\includegraphics[width=0.18\textwidth]{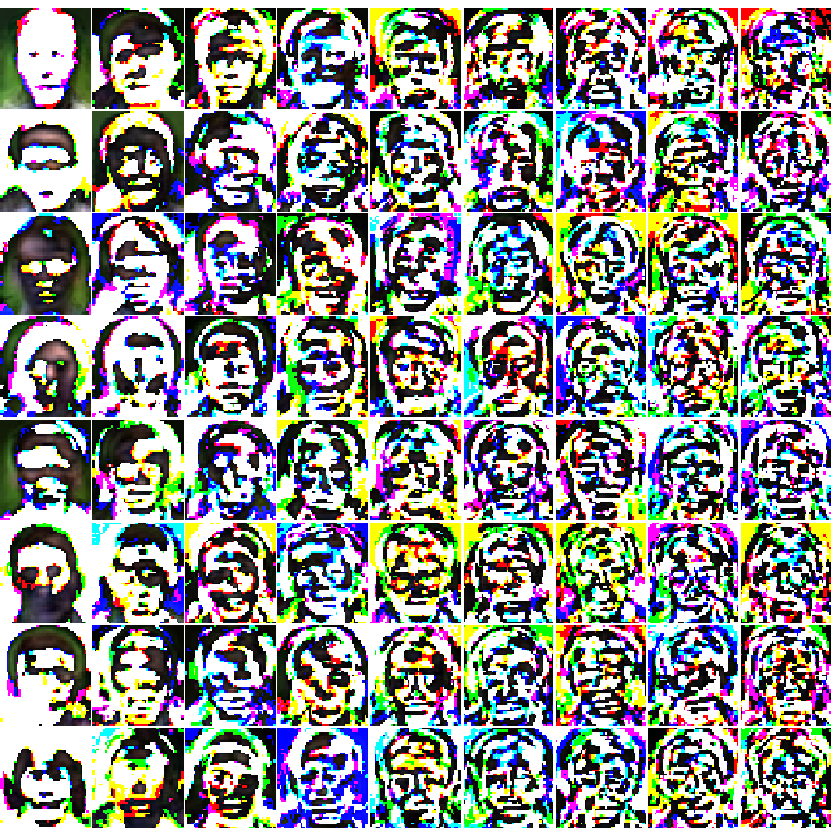}}
	\subfigure[ ]{
		\includegraphics[width=0.18\textwidth]{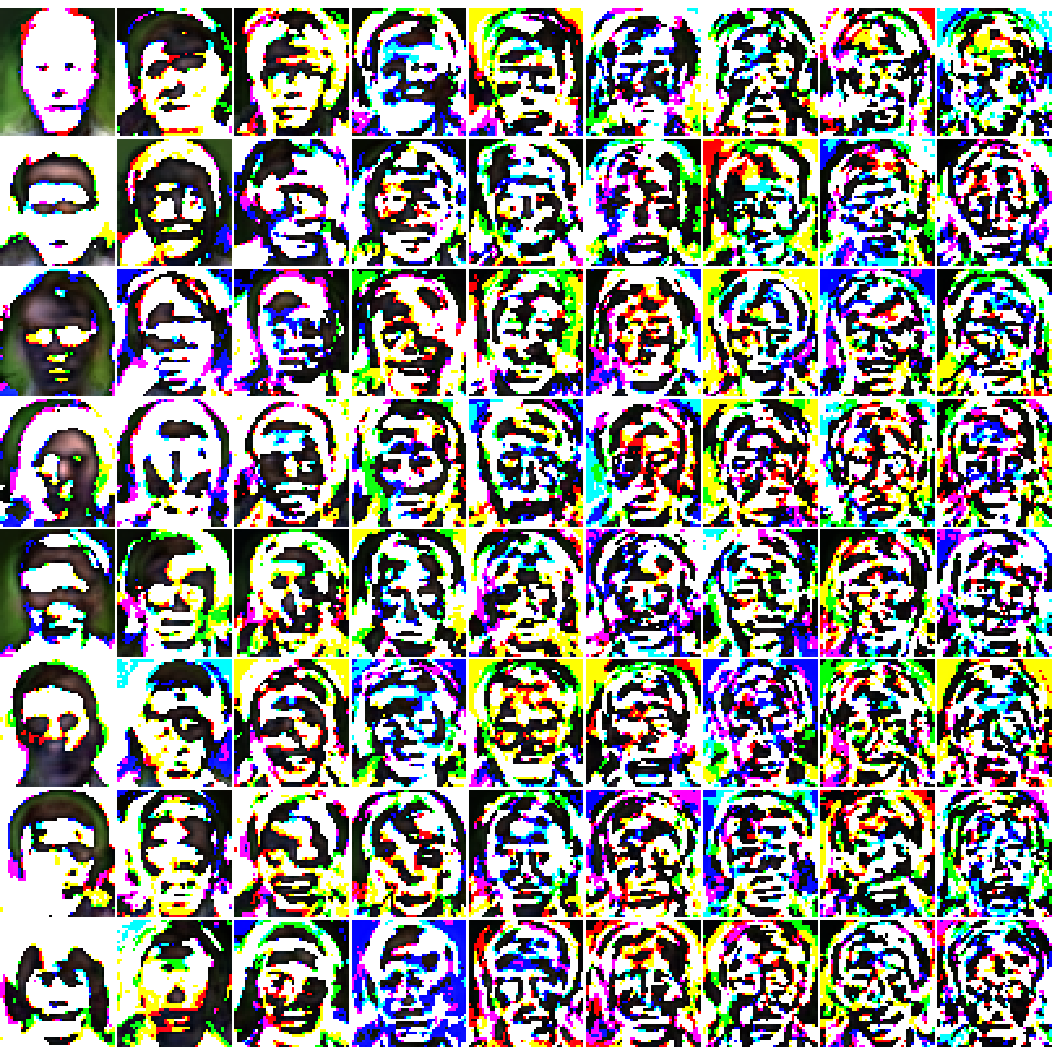}}
	\subfigure[ ]{
		\includegraphics[width=0.18\textwidth]{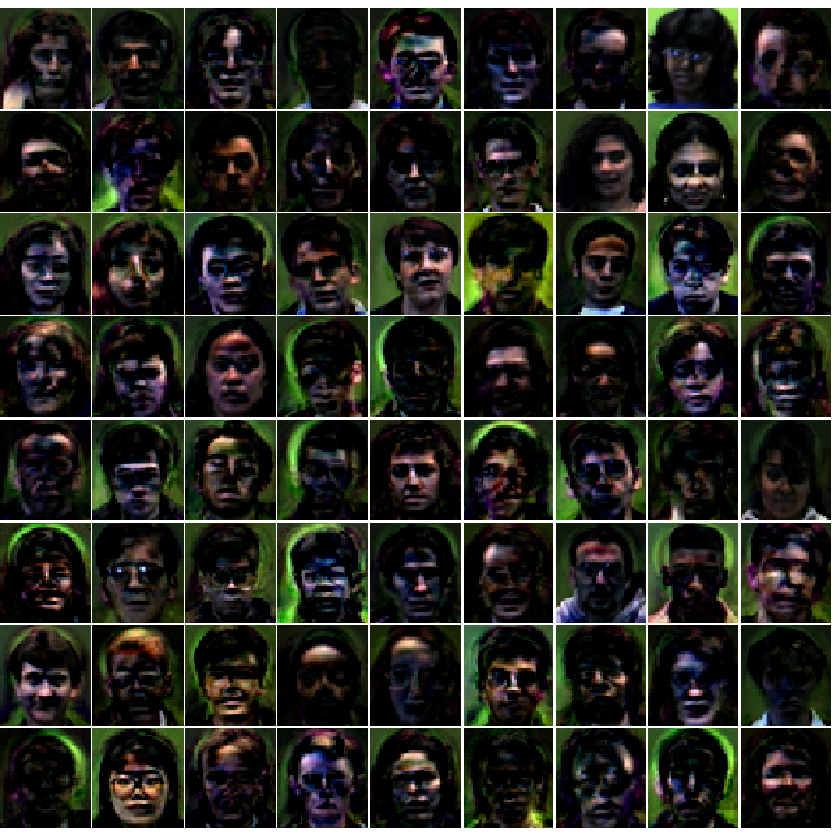}}
	\subfigure[ ]{
		\includegraphics[width=0.18\textwidth]{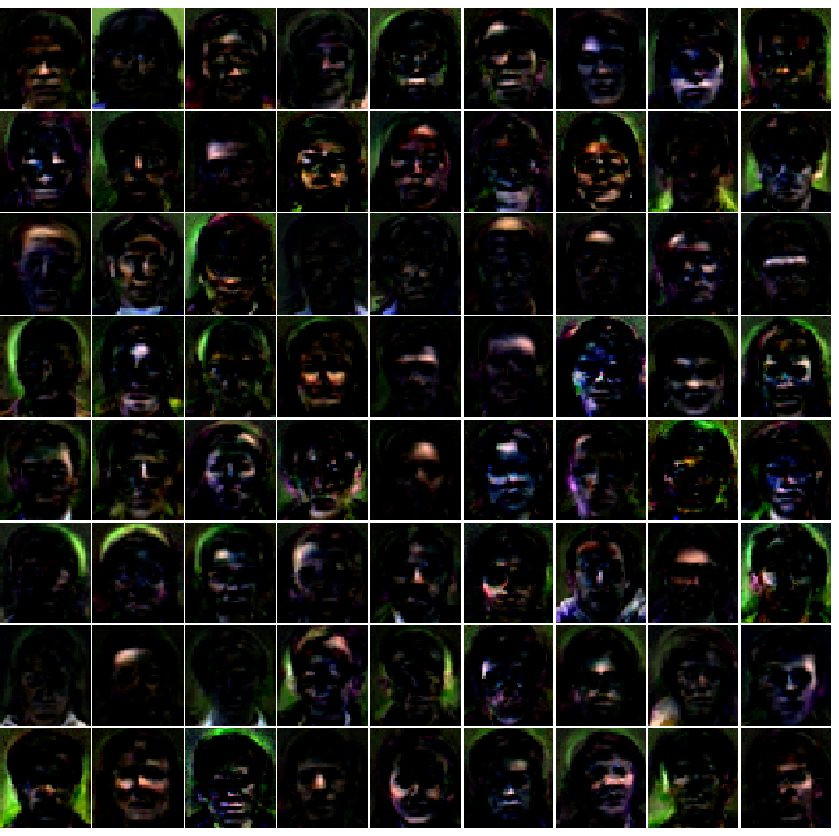}}\\
	\subfigure[ ]{
		\includegraphics[width=0.18\textwidth]{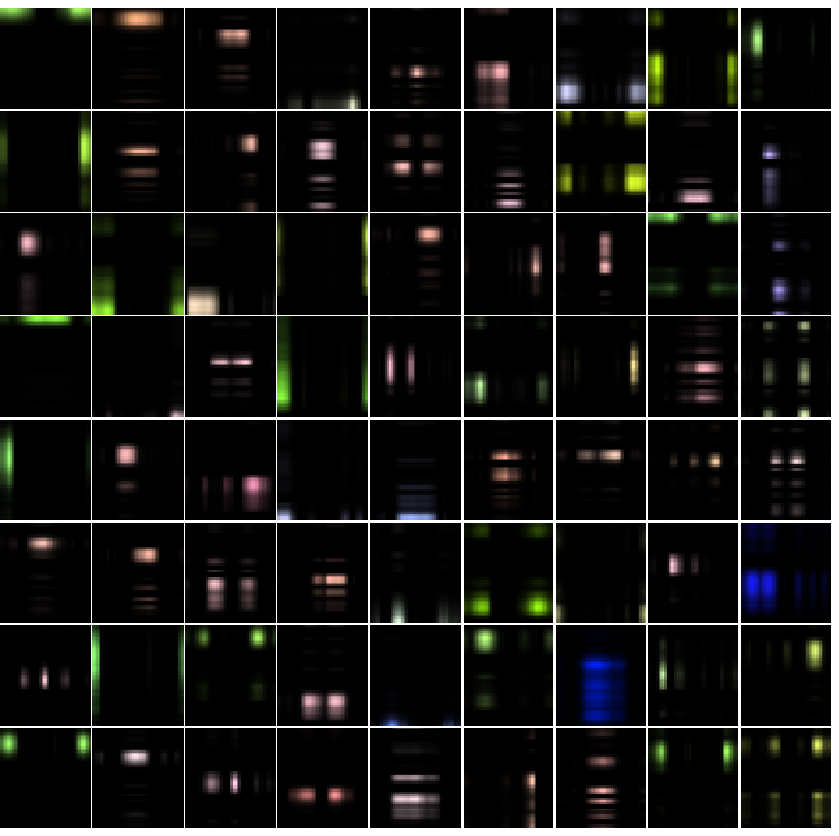}}
	\subfigure[ ]{
		\includegraphics[width=0.18\textwidth]{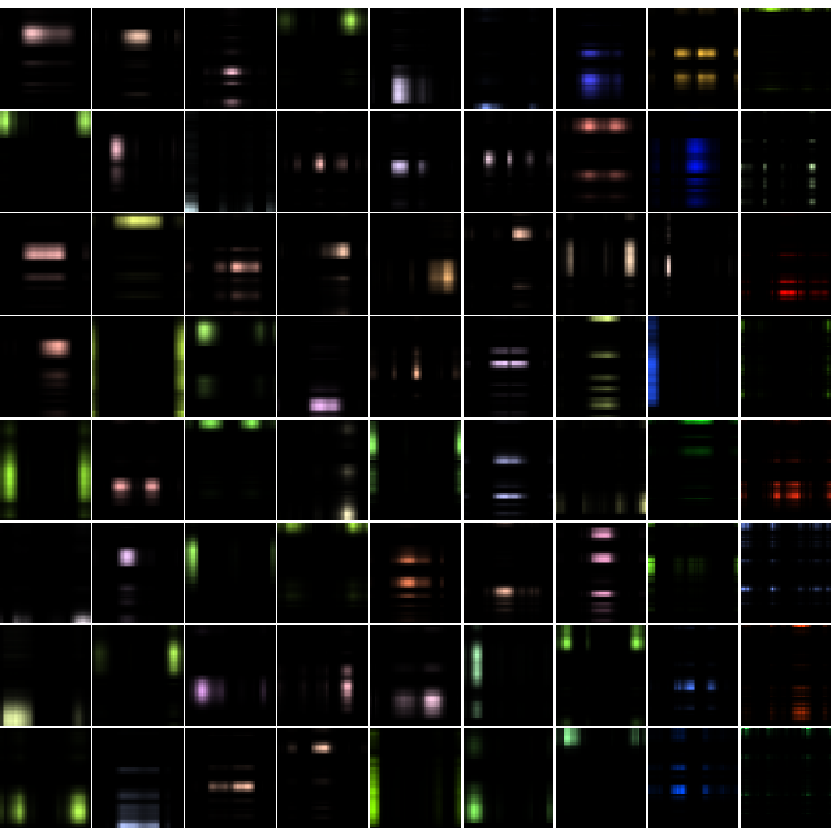}}
	\subfigure[ ]{
		\includegraphics[width=0.18\textwidth]{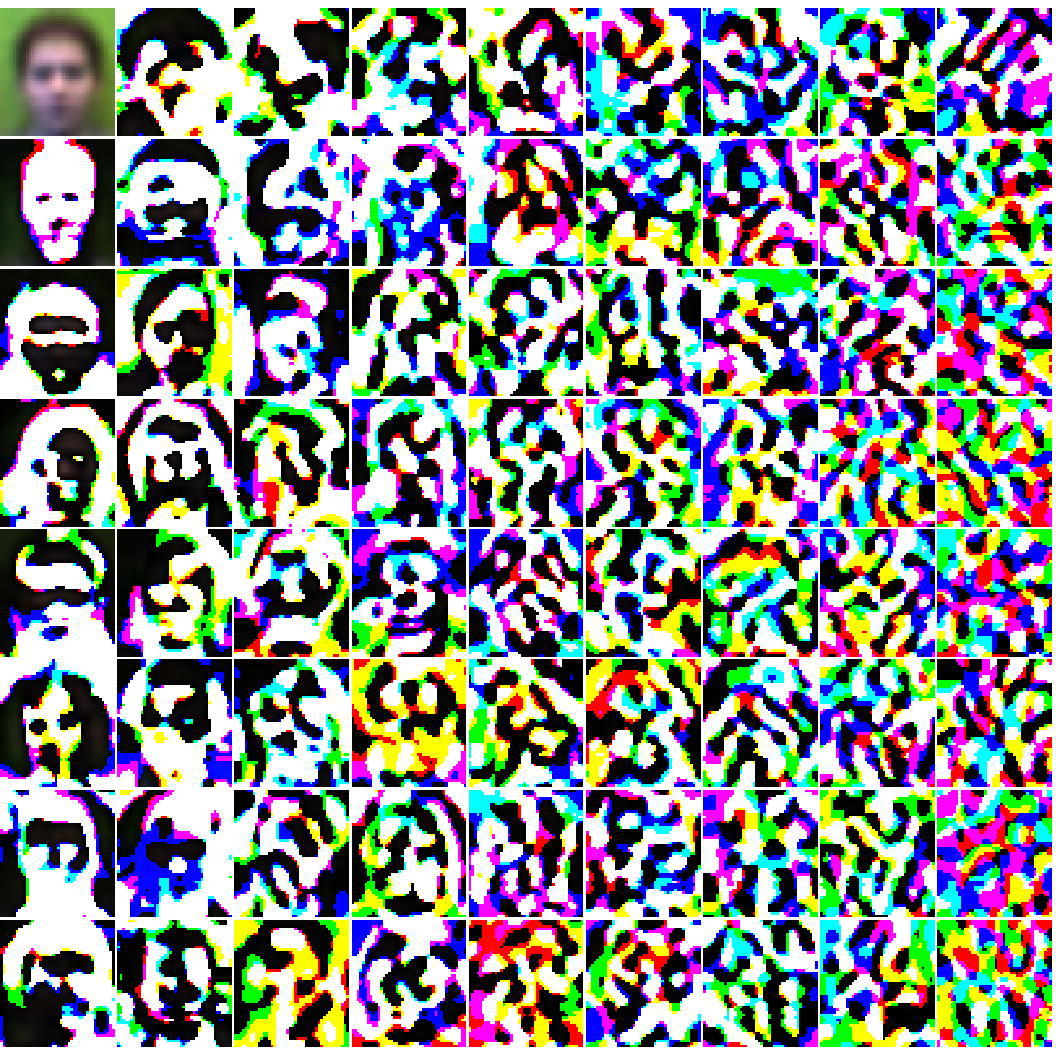}}
	\subfigure[ ]{
		\includegraphics[width=0.18\textwidth]{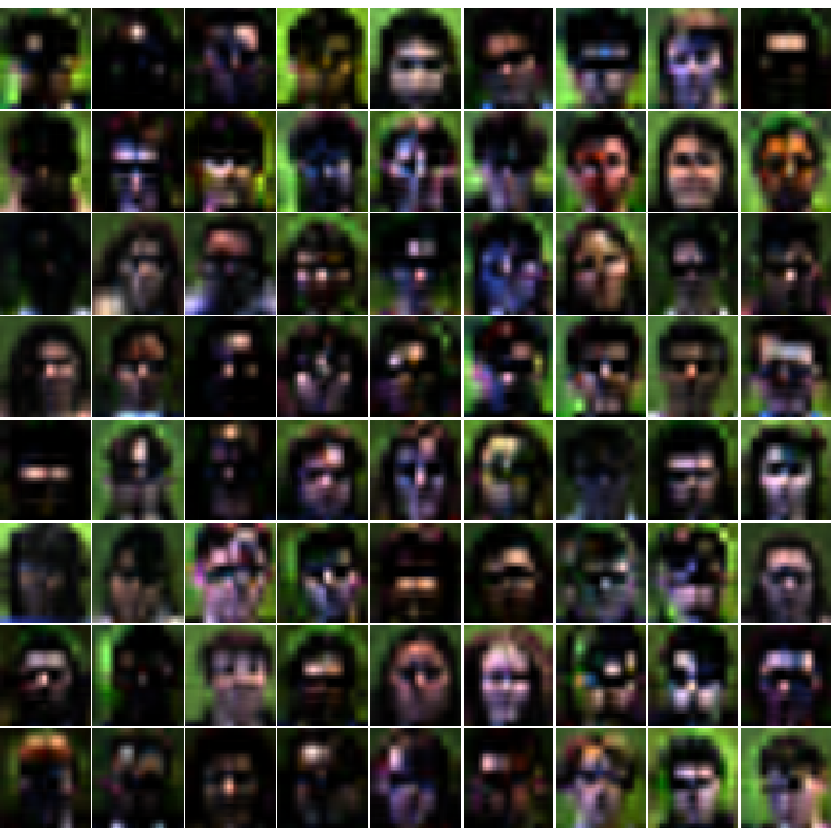}}
	\subfigure[ ]{
		\includegraphics[width=0.18\textwidth]{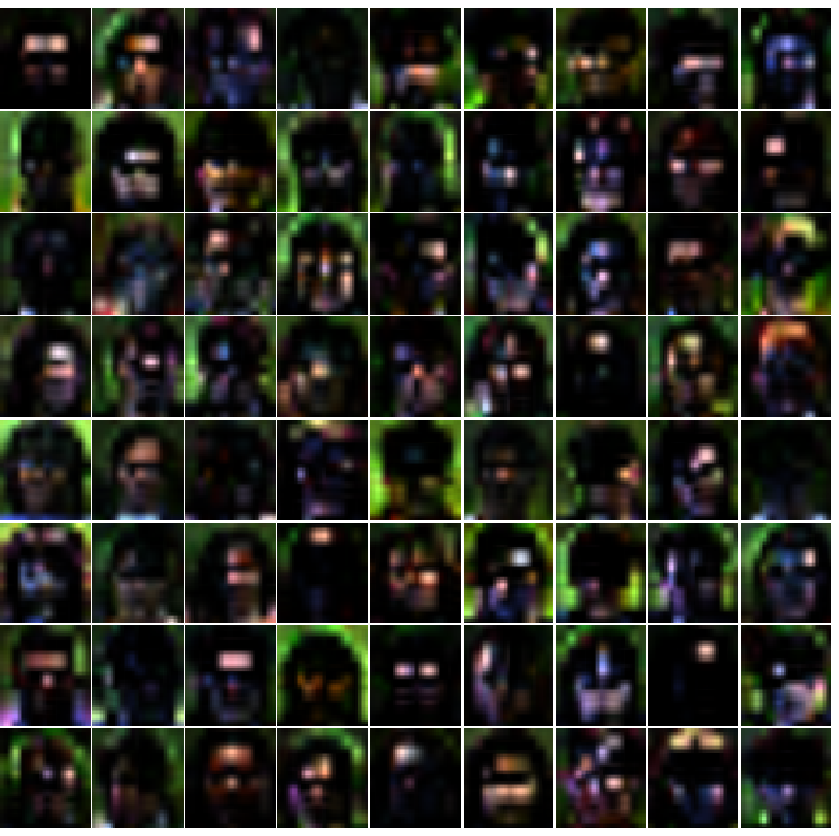}}\\
	\subfigure[ ]{
		\includegraphics[width=0.18\textwidth]{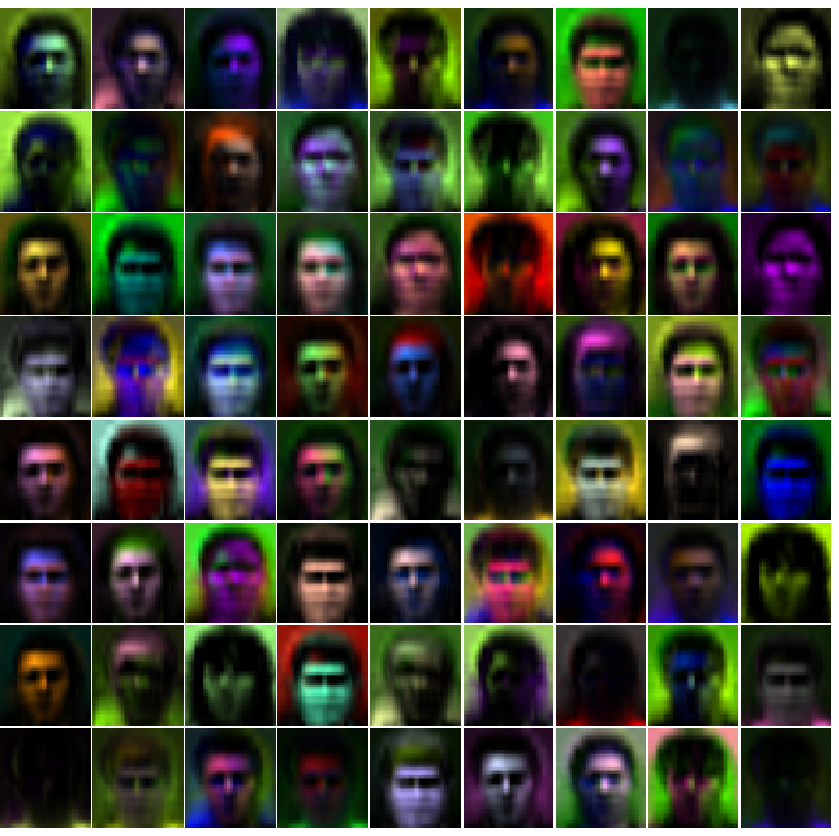}}
	\subfigure[ ]{
		\includegraphics[width=0.18\textwidth]{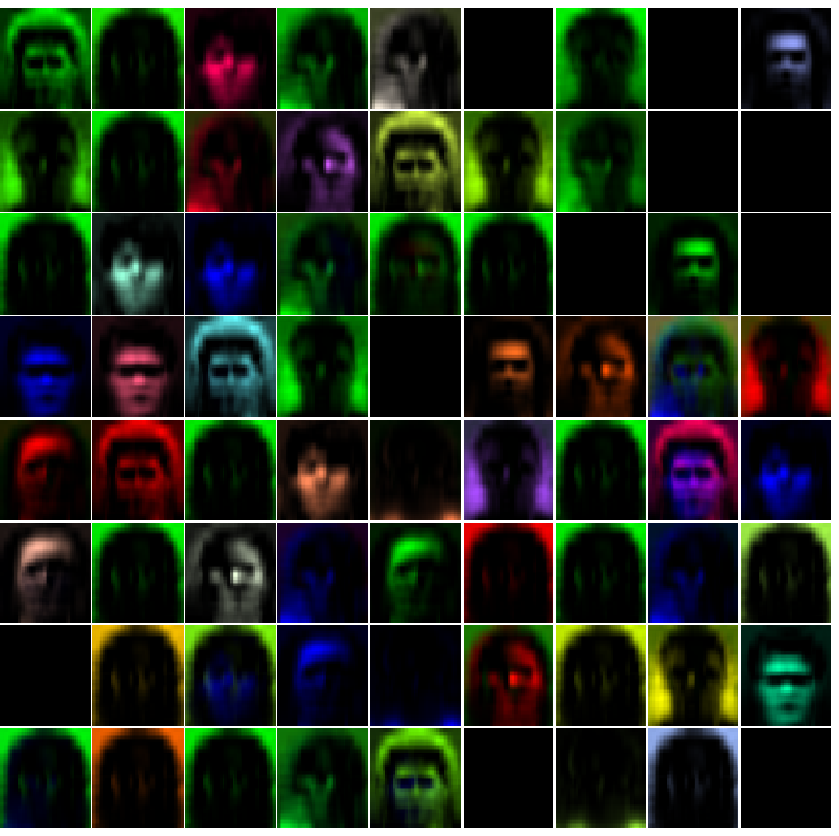}}
	\subfigure[ ]{
		\includegraphics[width=0.18\textwidth]{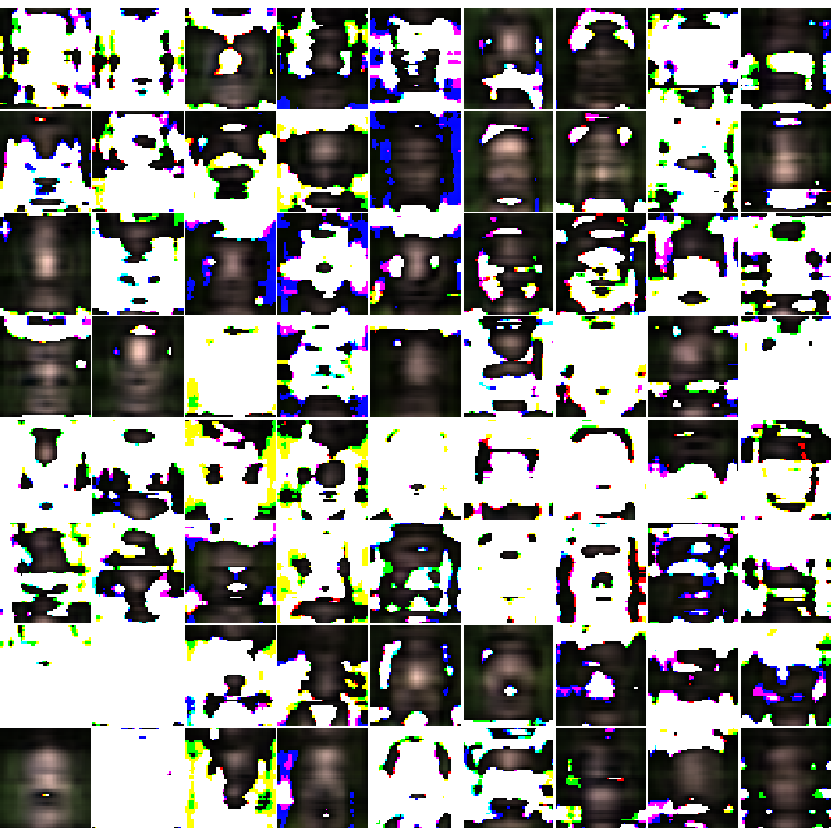}}
	\subfigure[ ]{
		\includegraphics[width=0.18\textwidth]{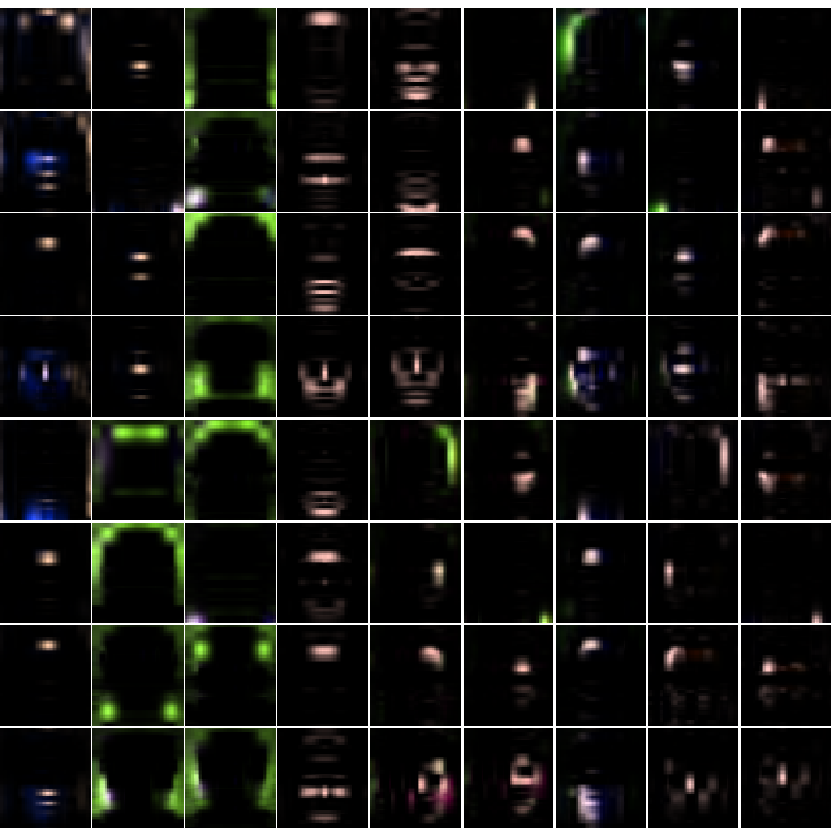}}
	\subfigure[ ]{
		\includegraphics[width=0.18\textwidth]{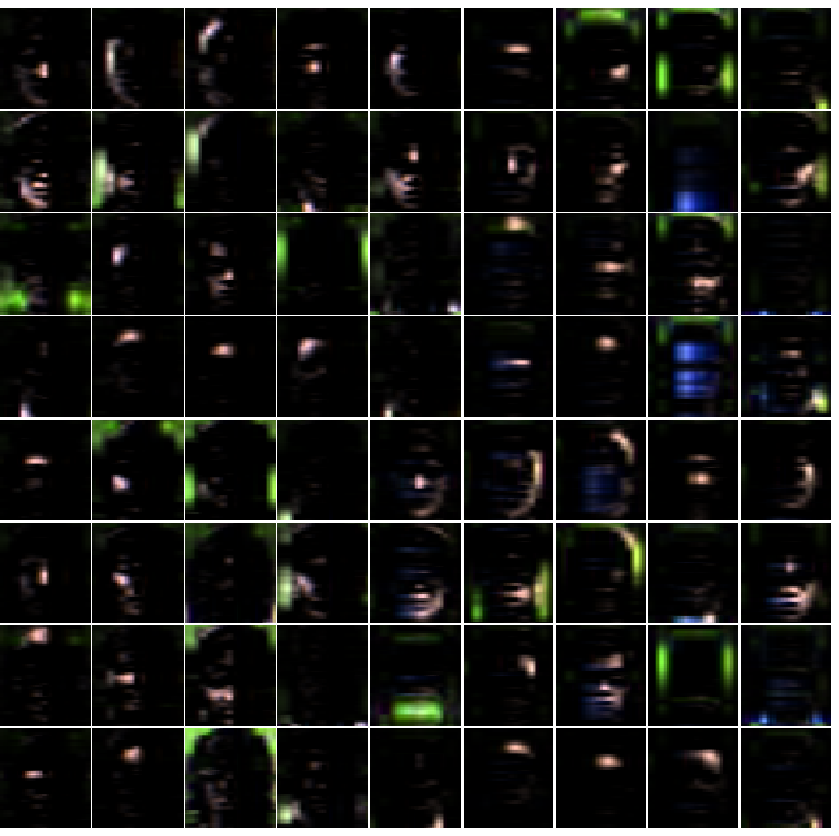}}\\
	\caption{Visualization of the Faces94 database. (a) Objects of the Faces94 database. The visualization of the basis extracted by different algorithms: (b) PCA. (c) gLPCA. (d) NMF. (e) GNMF. (f) NTF. (g) LRNTF. (h) GLTD. (i) NTD. (j) GNTD. (k) NTT. (l) GNTT. (m) TR. (n) NTR. (o) GNTR. Each image denotes all basis extracted by the above-mentioned algorithms on Faces94 database (visually display negative elements as white).}
	\label{fig:3}
\end{figure*}

\subsection{Evaluation Measures}
To compare the ability of different algorithms to learn localized parts of the tensor objects in the basis visualization task, the sparseness level metric is adopted in this paper \cite{hoyer2004non}, as shown in
\begin{equation}
	\operatorname{Sparseness}\left( \mathbf{H} \right)= \frac{\sqrt{n}-\left\|\operatorname{vec}(\mathbf{H})\right\|_{1} / \left\|\operatorname{vec}(\mathbf{H})\right\|_{2}}{\sqrt{n}-1}, 
\end{equation}
where $\mathbf{H}$ and $\operatorname{vec}(\mathbf{H})$ denote basis extracted by the algorithm and its vector form, $n$ is denoted by the number of elements of $\mathbf{H}$. The $\left\|\cdot\right\|_{1}$ and $\left\|\cdot\right\|_{2}$ are defined by the $L1$ norm and $L2$ norm.

To quantitatively evaluate the effectiveness of each algorithm, we adopted two metrics of Accuracy (AC) and Normalized Mutual Information (NMI). The definition of AC is as follows
\begin{equation}
	\operatorname{AC}\left(y_{i}, \hat{y}_{i}\right)=\frac{1}{n} \sum_{i=1}^{n} \delta\left(y_{i}, \operatorname{map}\left(\hat{y}_{i}\right)\right), 
\end{equation}
where $n$ is the total number of objects. $y_{i}$ and $\hat{y}_{i}$ represent the cluster label of the object and the true label of the object. $\operatorname{map}\left( \cdot \right)$ denotes a displacement mapping function, which is responsible for mapping each cluster label $y_{i}$ to the equivalent label from the data corpus. If the object label $y_{i}$ and the real label $\hat{y}_{i}$ are equal, then $\left(y_{i}, \operatorname{map}\left(\hat{y}_{i}\right)\right)=1$, if not, then $\left(y_{i}, \operatorname{map}\left(\hat{y}_{i}\right)\right)=0$.

By employing the information theory, the agreement between two cluster partitions can be measured with mutual information (MI). The MI between the collection of cluster labels $C^{\prime}$ and the collection of true labels $C$ is defined by
\begin{equation}
	\operatorname{MI}\left(C, C^{\prime}\right)=\sum_{c_{i} \in C, c_{i}^{\prime} \in C^{\prime}} p\left(c_{i}, c_{i}^{\prime}\right) \cdot \log _{2} \frac{p\left(c_{i}, c_{i}^{\prime}\right)}{p\left(c_{i}\right) \cdot p\left(c_{i}^{\prime}\right)}, 
\end{equation}
where $p\left(c_{i}\right)$ and $p\left(c_{i}^{\prime}\right)$ denote the object belongs to the probability of category $c_{i}$ and category $c_{i}^{\prime}$ that the random selection of a object from the databases. $p\left(c_{i}, c_{i}^{\prime}\right)$ is defined by the object belongs to the probability of category $c_{i}$ and category $c_{i}^{\prime}$ as the same time that the random selection of a object from the databases. To force the score to have an upper bound, we used the NMI as one of evaluation measures and the definition of NMI is denoted as follows
\begin{equation}
	\operatorname{NMI}\left(C, C^{\prime}\right)=\frac{\operatorname{MI}\left(C, C^{\prime}\right)}{\max \left(H(C), H\left(C^{\prime}\right)\right)}, 
\end{equation}
where $H\left(C\right)$ and $H\left(C^{\prime}\right)$ are defined as the entropy of the true label collection $C$ and the entropy of the cluster label collection $C^{\prime}$. It is quite straightforward to know the score ranges of $\operatorname{NMI}\left(C, C^{\prime}\right)$ from $0$ to $1$, with $\operatorname{NMI}\left(C, C^{\prime}\right)=1$ if the two label collection are the same, and $\operatorname{NMI}\left(C, C^{\prime}\right)=0$ otherwise.

\begin{figure*}[!t]
	\centering  
	\subfigure[ ]{
		\includegraphics[width=0.458\textwidth]{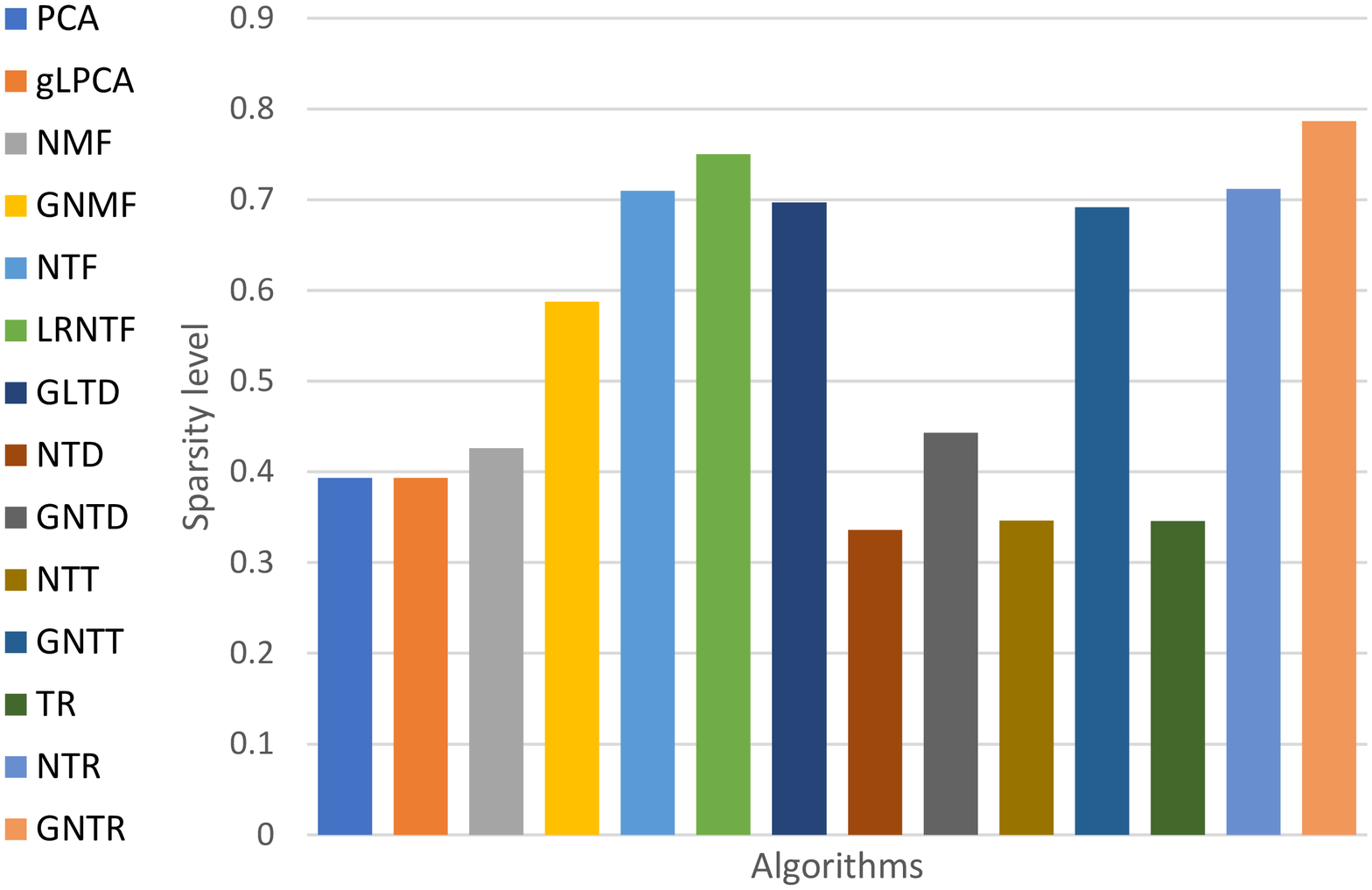}}
	\hspace{0.2cm}
	\subfigure[ ]{
		\includegraphics[width=0.40\textwidth]{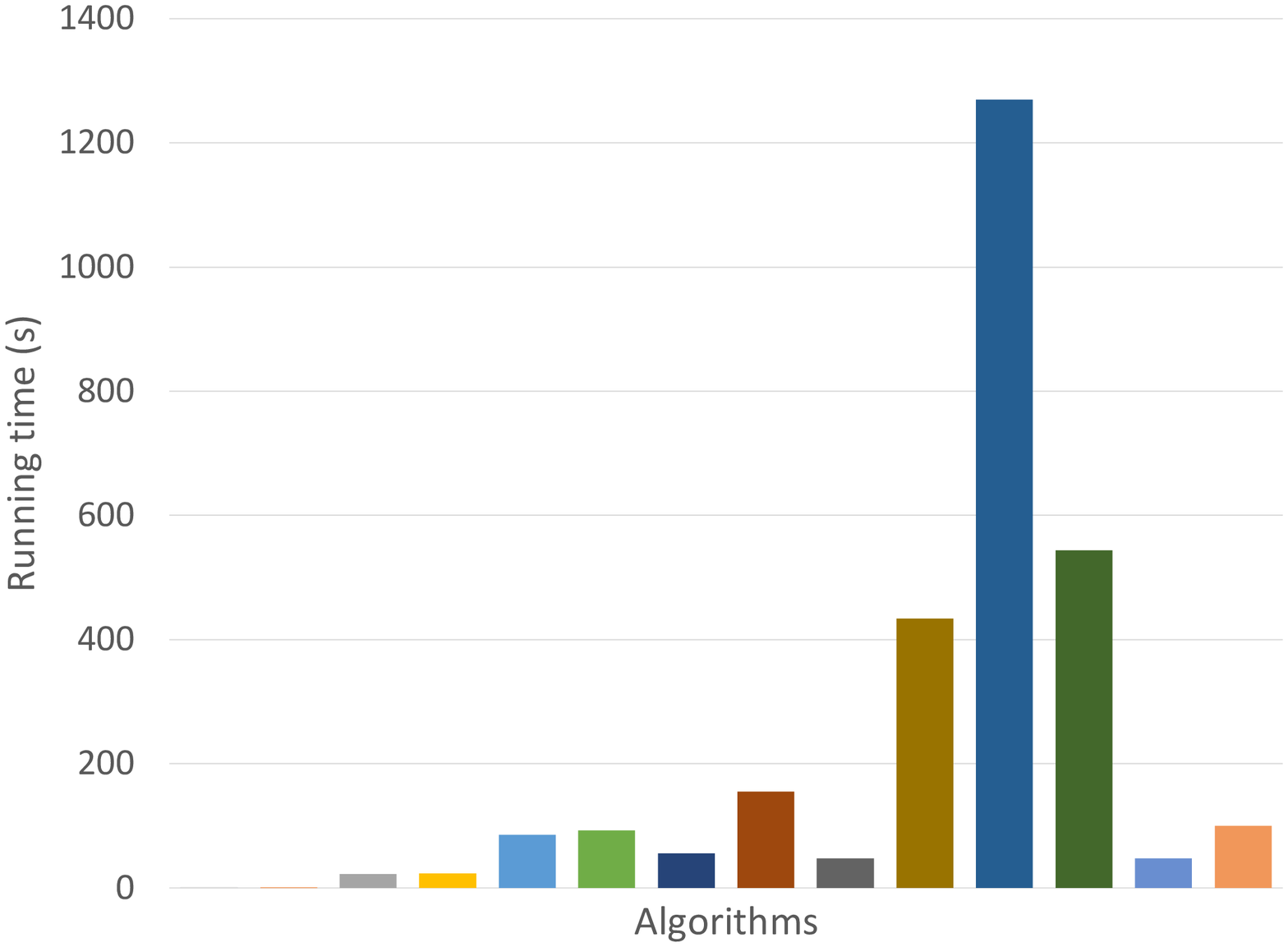}} 
	
	\caption{Comparison between different algorithms in particular settings on the Faces94 database. (a) Sparsity level. (b) Running time.}
	\label{fig:4}
\end{figure*}

\begin{figure*}[!t]
	\centering  
	\subfigure[ ]{
		\includegraphics[width=0.15\textwidth]{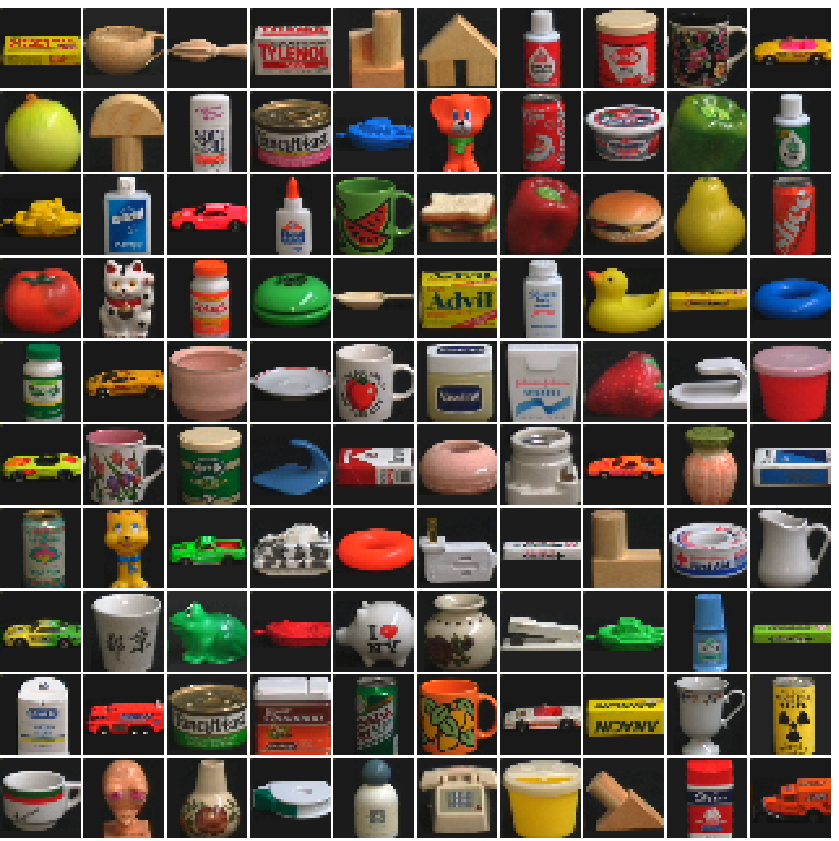}}
	\subfigure[ ]{
		\includegraphics[width=0.15\textwidth]{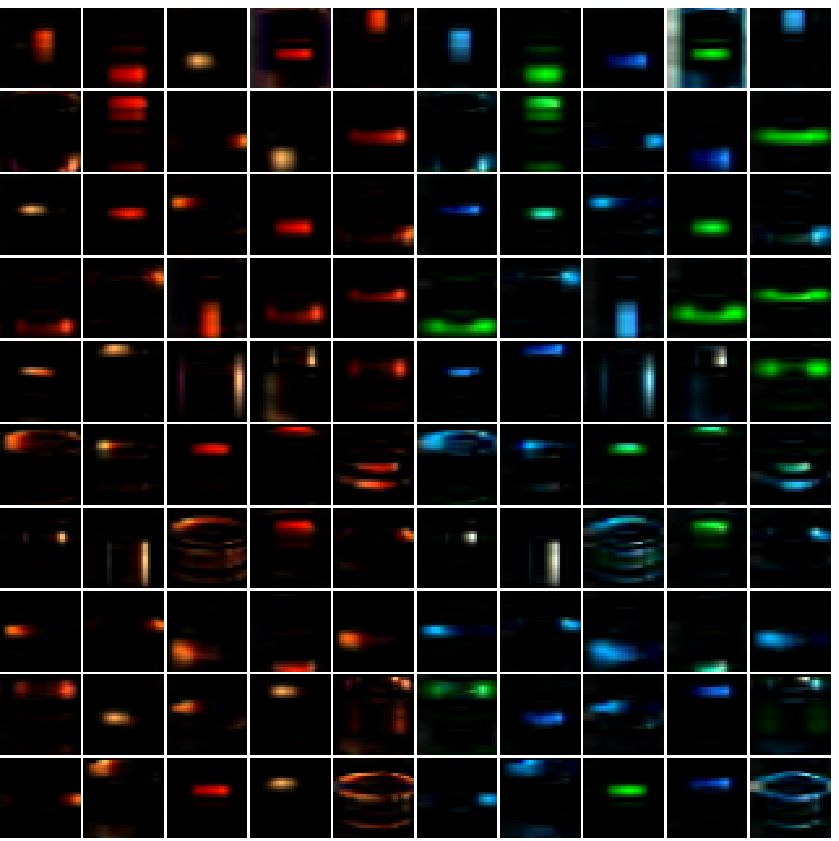}}
	\subfigure[ ]{
		\includegraphics[width=0.15\textwidth]{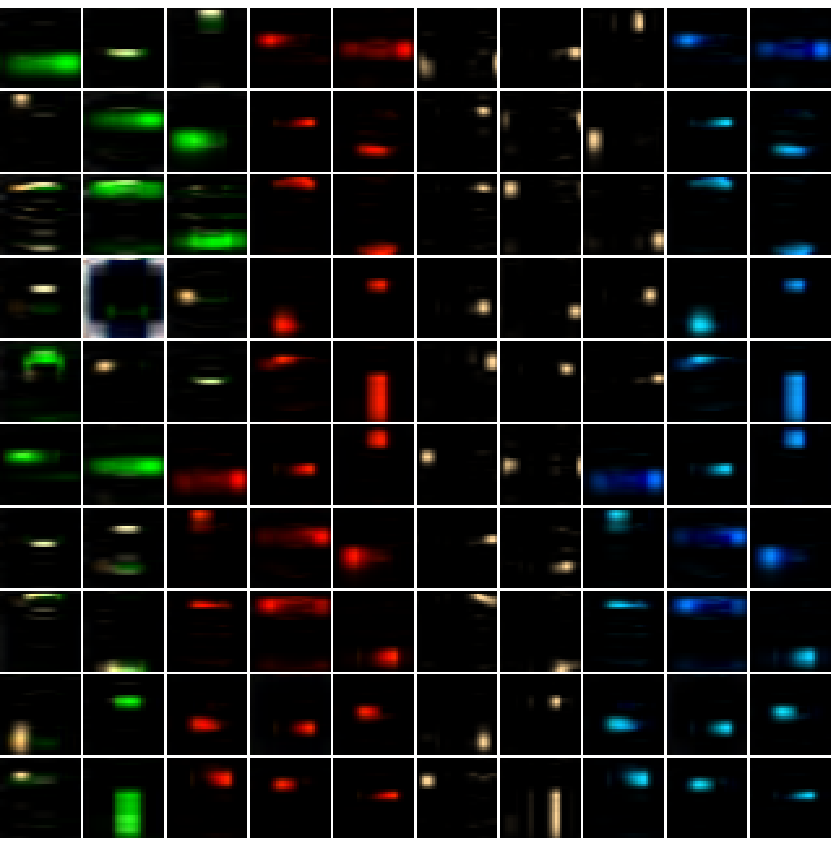}}
	\subfigure[ ]{
		\includegraphics[width=0.15\textwidth]{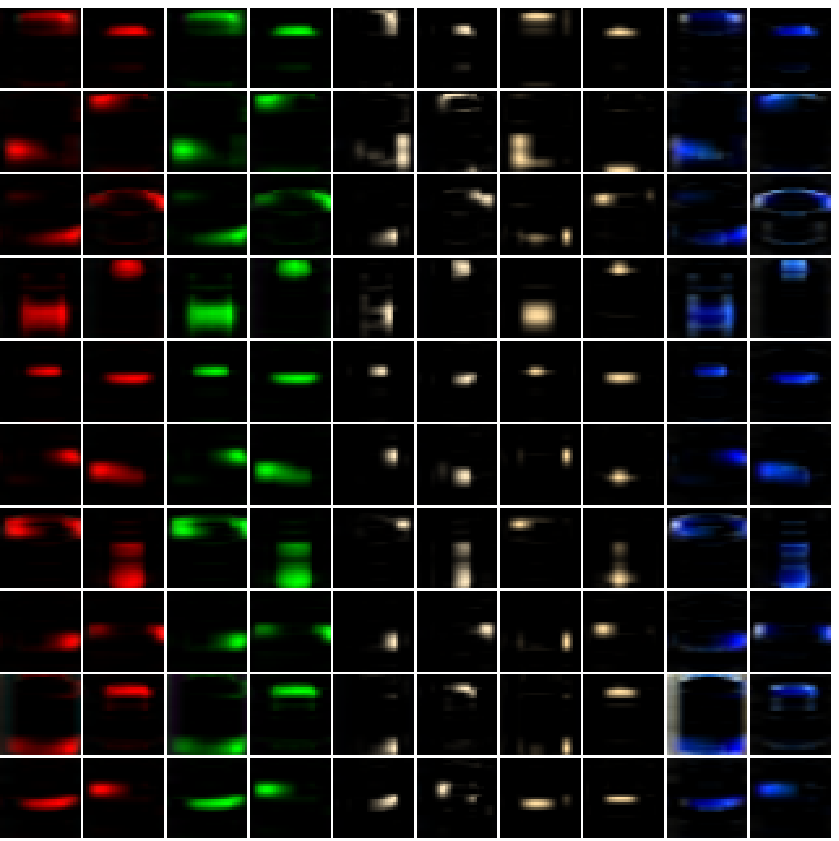}}
	\subfigure[ ]{
		\includegraphics[width=0.15\textwidth]{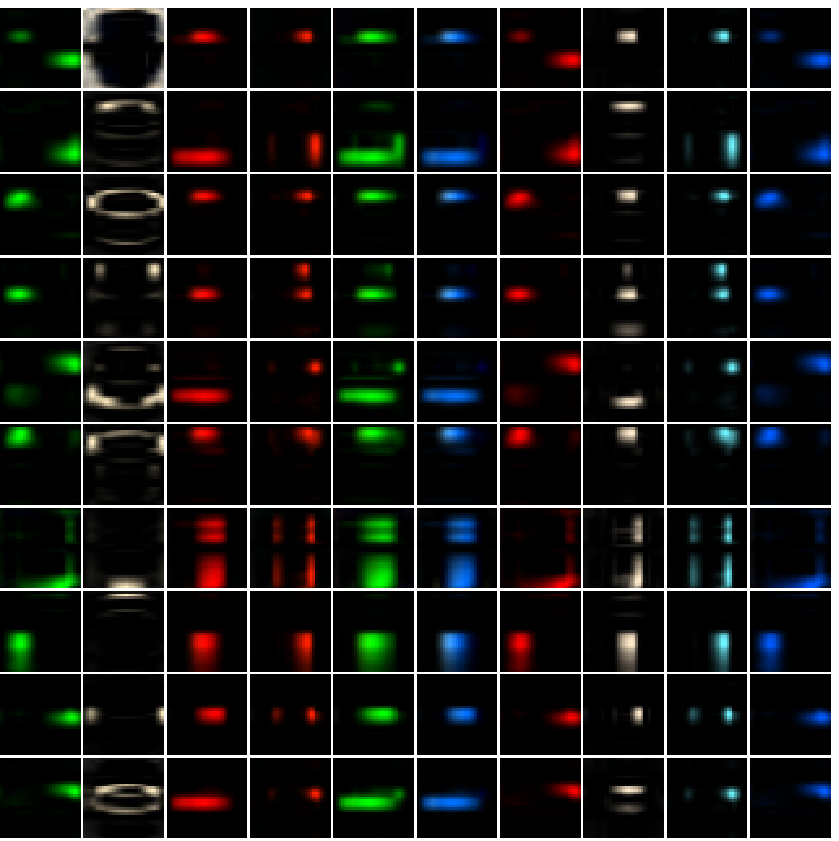}}
	\subfigure[ ]{
		\includegraphics[width=0.15\textwidth]{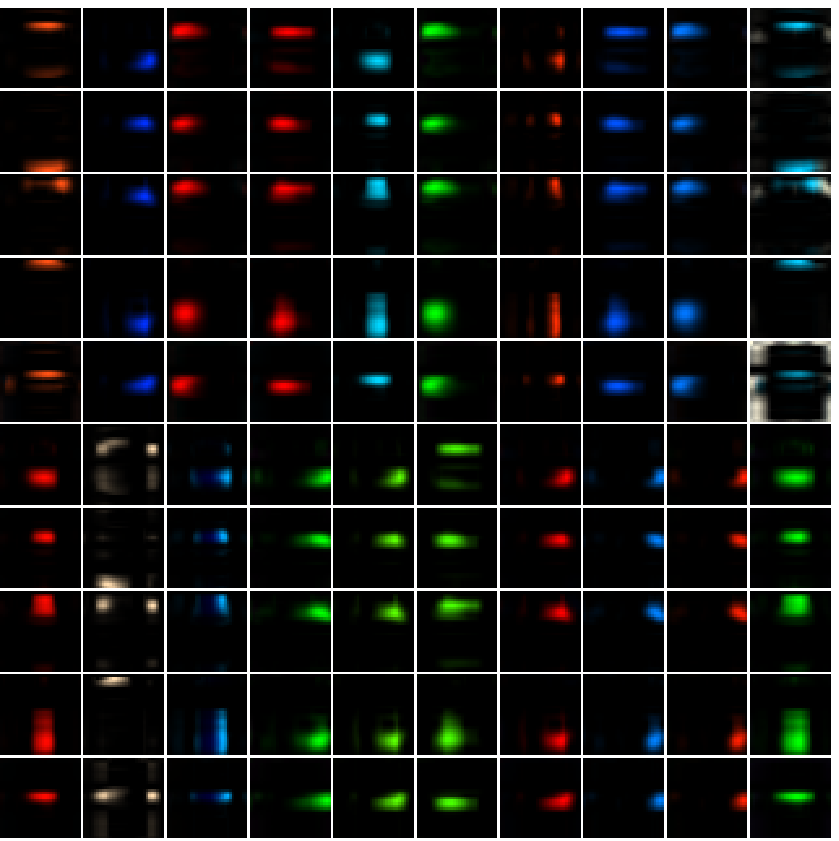}}
	\caption{The visualization of the objects of COIL-100 database and the visualization of basis extracted of NTR algorithm under different combinations of nonnegative multiway rank of the feature core tensor. (a) The visualization of the objects of COIL-100 database. The different size of the feature core tensor $\mathcal{G}^{(4)}$: (b) $\mathcal{G}^{(4)} \in \mathcal{R}^{2 \times 100 \times 50}$. (c) $\mathcal{G}^{(4)} \in \mathcal{R}^{4 \times 100 \times 25}$. (d) $\mathcal{G}^{(4)} \in \mathcal{R}^{5 \times 100 \times 20}$. (e) $\mathcal{G}^{(4)} \in \mathcal{R}^{10 \times 100 \times 10}$. (f) $\mathcal{G}^{(4)} \in \mathcal{R}^{20 \times 100 \times 5}$.}
	\label{fig:5}
\end{figure*}

\subsection{Basis Visualization}
The nonnegativity of the data representation brings about two key effects: the data representation is purely additive, and it can provide more interpretable and meaningful representation for physical signals. The data representation are often sparse because they often contain many zero entries. Due to these two effects, the nonnegative decomposition algorithms have the ability to extract the parts-based basis of tensor objects.

To compare the ability to extract the parts-based basis of tensor objects by each algorithms, we visual the basis extracted by each algorithms in Faces94 database. The number of basis is set to the number of categories $k$ of the Faces94 database, that is, 72 basis images are extracted, and stitched them together, as shown in Fig. \ref{fig:3}. We also count the sparsity level and running times of each algorithms and display them in Fig. \ref{fig:4}. The experimental results are summarized as follows:

\begin{itemize}
	\item The proposed NTR and GNTR algorithms extract the parts-based basis with rich colors and rich lines from the Faces94 database, which can provide more interpretable and meaningful representation for tensor objects. The basis extracted by TR is less sparse, and it has been shown that nonnegative decomposition algorithm can provide a more sparse representation than unconstrained algorithm, which is consistent with results \cite{hazan2005sparse} \cite{qiu2020generalized}.
	\item NTF and LRNTF extract the sparse basis of the tensor objects by considering the high-dimensional structures in image space, which is consistent with results \cite{hazan2005sparse} \cite{wang2011image}. However, the lines of the basis extracted by the NTF model are too abstract to identify the outline of the tensor object. The reason possibly is NTF and LRNTF assume that the low-rank properties of different dimensions of tensor data are the same. The sparse basis extracted by the proposed NTR and GNTR is rich lines that can identify the outline of the tensor object due to these algorithms can capture the low-rank properties between each dimension by nonnegative multiway rank.
	\item NTT and GNTT extract the basis with rich colors and poor lines from Faces94 database. The reason is that the feature core tensor can only maintain direct connection and interaction with the third core tensor that stores color information, not the first core tensor that stores row pixels information. NTR and GNTR inherited the circular dimensional permutation invariance from TR decomposition, and thus the feature core tensor of these models can directly maintain connection and interaction with the first core tensor. Therefore, NTR and GNTR can extract the parts-based basis with rich colors and rich lines from the Faces94 database.
	\item Our proposed NTR and GNTR algorithms are faster than the TR algorithm based on alternating least squares method, which is shown that the efficiency of the APG method. The third core tensor of the NTT model shares a nonnegative multiway rank with the feature core tensor representing the feature, and this rank is set to the number of categories k of the Faces94 database in our task. Therefore, the NTT and GNTT algorithms need to manipulating huge matrix when updating the third core tensor, and it is particularly time-consuming. This problem can reduce the impact by permuting the dimensions of the data, and there are many ways to permute. To keep the tensor data is same structure as the data used by the other algorithms and facilitate the task, we do not consider to permute the dimensions of tensor data in the task of NTT and GNTT algorithm.
\end{itemize}

\begin{table*}[!t]
	\renewcommand{\arraystretch}{1.2}
	\caption{AC and NMI of different algorithms on five databases. The best results are marked by bold font and the second best results are marked by underline.}
	\centering
	\setlength{\tabcolsep}{1mm}
	\begin{tabular}{c|c|ccccccccccccccc}
		\hline
		\hline
		Algorithms & Metric & Original & PCA & gLPCA & NMF & GNMF & NTF & LRNTF & GLTD & NTD & GNTD & NTT & GNTT & TR & NTR & GNTR\\ \hline
		
		\multirow{2}{*}{ORL} & AC & 67.0 & 61.2 & 66.1 & 67.9 & 73.3 & 66.9 & \underline{74.4} & 63.2 & 67.6 & 73.6 & 53.0 & 69.1 & 48.4 & 66.7 & \textbf{75.8}\\
		
		& NMI & 83.2 & 78.9 & 80.2 & 82.9 & \underline{86.8} & 81.2 & 87.2 & 78.8 & 82.2 & 86.3 & 72.5 & 83.7 & 67.6 & 82.0 & \textbf{87.8}\\ \hline
		
		\multirow{2}{*}{FEI PART 1} & AC & 54.4 & 55.3 & 54.7 & 56.8 & 63.4 & 51.8 & 64.3 & 51.8 & 60.1 & 68.1 & 33.4 & 61.3 & 43.0 & \underline{69.7} & \textbf{73.4}\\
		
		& NMI & 75.6 & 75.0 & 72.6 & 75.9 & 80.5 & 73.5 & 81.8 & 61.5 & 77.4 & 83.5 & 55.4 & 78.2 & 65.0 & \underline{84.4} & \textbf{86.6}\\ \hline
		
		\multirow{2}{*}{GT} & AC & 44.9 & 41.6 & 41.7 & 45.5 & 44.0 & 42.1 & \underline{47.5} & 40.9 & 43.5 & 40.3 & 42.1 & 40.3 & 30.1 & 47.4 & \textbf{51.9}\\ 
		
		& NMI & 62.5 & 59.1 & 59.6 & 63.1 & 62.2 & 60.8 & 64.9 & 59.8 & 61.8 & 59.3 & 63.3 & 60.8 & 50.9 & \underline{65.5} & \textbf{68.5}\\ \hline
		
		\multirow{2}{*}{COIL-100 PART 1} & AC & 70.4 & 70.0 & 70.0 & 69.5 & 74.6 & 69.1 & 76.4 & 70.6 & 73.5 & \underline{77.1} & 75.8 & 80.0 & 70.4 & 72.9 & \textbf{84.2}\\
		
		& NMI & 80.2 & 79.4 & 79.6 & 80.2 & 85.0 & 78.8 & 84.6 & 80.2 & 79.3 & 86.2 & 83.7 & \underline{88.8} & 78.7 & 81.5 & \textbf{90.5}\\ \hline
		
		\multirow{2}{*}{Faces94 PART 1} & AC & 74.4 & 71.6 & 71.7 & 77.1 & 76.5 & 77.6 & 75.7 & 73.2 & 76.9 & 75.8 & 75.8 & 74.1 & \textbf{80.8} & \underline{80.5} & 77.6\\
		
		& NMI & 90.9 & 88.6 & 88.4 & 91.3 & 92.1 & 91.5 & 91.9 & 89.7 & 91.2 & 92.2 & 89.7 & 91.5 & 92.6 & \textbf{92.9} & \textbf{92.9}\\
		\hline
		\hline
	\end{tabular}
	\label{table2}
\end{table*}

We conducted basis visualization tasks on the COIL-100 data with NTR under different combinations of nonnegative multiway rank of the feature core tensor to verify the ability to extract the parts-based basis of tensor objects. Fig. \ref{fig:5} shown that the visualization of basis extracted of the proposed NTR under the different nonnegative multiway rank settings of the feature core tensor. The task results has brought about a discovery of some rule, as the nonnegative multiway rank shared with the feature core tensor and the third core tensor that stores color information increases, the color richness of the basis visualization results increases. As the nonnegative multiway rank shared with the feature core tensor and the first core tensor that stores row pixels information increases increases, the line richness of the basis visualization results increases. This is an interesting phenomenon, as far as we know, it has never been observed in previous studies of the same-type algorithms. The reason is that NTR inherits the circular dimensional permutation invariance from TR \cite{zhao2019learning}, the feature core tensor can maintain directly connection and interaction with adjacent core tensors. Therefore, we can obtain the basis representations with rich colors and rich lines by control the rank combination of the feature core tensor, which is provide more interpretable and meaningful representation.

In practical applications, for the tensor objects with rich colors, we can choose the nonnegative multiway rank combination that highlights the color richness of basis to enhance the effectiveness of data representation. For the tensor objects with rich lines, we can choose the nonnegative multiway rank combination that highlights the line richness of basis to better depict the details of different objects.

\begin{table*}[!t]
	\renewcommand{\arraystretch}{1.2}
	\caption{Classification results (40\% labeled data) using k-NN classification algorithms. The best results are marked by bold font and the second best results are marked by underline.}
	\centering
	\setlength{\tabcolsep}{1mm}
	\begin{tabular}{c|c|ccccccccccccccc}
		\hline
		\hline
		Algorithms & Metric & Original & PCA & gLPCA & NMF & GNMF & NTF & LRNTF & GLTD & NTD & GNTD & NTT & GNTT & TR & NTR & GNTR\\ \hline
		
		\multirow{3}{*}{ORL} & k-NN(k=1) & 89.6 & 86.7 & 84.2 & \underline{89.8} & 89.5 & 87.3 & 89.6 & 88.8 & 80.0 & 73.3 & 88.6 & 87.8 & 72.0 & 87.6 & \textbf{90.2}\\
		
		& k-NN(k=3) & 77.5 & 71.7 & 72.9 & 78.9 & 80.9 & 75.8 & \underline{81.8} & 77.3 & 73.3 & 79.8 & 58.3 & 79.8 & 57.0 & 77.5 & \textbf{82.1}\\
		
		& k-NN(k=5) & 70.4 & 70.4 & 72.5 & 73.8 & 75.8 & 70.5 & \underline{77.6} & 68.8 & 74.2 & 74.7 & 55.1 & 72.4 & 52.8 & 71.0 & \underline{77.5}\\ \hline
		
		\multirow{3}{*}{FEI PART 1} & k-NN(k=1) & 80.2 & 69.8 & 67.6 & 68.6 & 75.6 & 68.3 & 76.0 & 70.0 & 66.1 & 75.1 & 26.8 & 73.9 & 63.7 & \textbf{88.9} & \underline{83.6}\\
		
		& k-NN(k=3) & 77.6 & 68.2 & 64.9 & 66.5 & 72.6 & 65.6 & 73.2 & 68.4 & 62.3 & 75.4 & 26.0 & 71.9 & 60.7 & \textbf{84.8} & \underline{81.6}\\
		
		& k-NN(k=5) & 70.7 & 63.3 & 62.4 & 65.4 & 72.3 & 63.9 & 74.3 & 63.1 & 61.8 & 76.5 & 26.8 & 71.1 & 57.4 & \textbf{84.1} & \underline{79.1}\\ \hline
		
		\multirow{3}{*}{GT} & k-NN(k=1) & \underline{62.2} & 56.4 & 56.9 & 58.0 & 54.2 & 58.2 & 58.2 & 58.7 & 57.3 & 49.6 & 42.6 & 51.8 & 43.3 & \textbf{63.8} & 59.9\\ 
		
		& k-NN(k=3) & 53.8 & 43.8 & 45.8 & 49.0 & 49.6 & 49.9 & 54.1 & 50.4 & 49.5 & 46.4 & 40.7 & 48.0 & 36.0 & \textbf{58.1} & \underline{54.8}\\
		
		& k-NN(k=5) & 53.1 & 41.8 & 43.1 & 49.3 & 48.4 & 49.6 & 54.3 & 49.8 & 49.1 & 46.5 & 40.5 & 47.2 & 35.8 & \textbf{57.7} & \underline{55.4}\\ \hline
		
		\multirow{3}{*}{COIL-100 PART 1} & k-NN(k=1) & 92.7 & 90.3 & 90.3 & 89.9 & 94.2 & 91.3 & 93.7 & 90.7 & 89.9 & 93.0 & 94.5 & \textbf{97.5} & 89.7 & 94.0 & \underline{95.8}\\
		
		& k-NN(k=3) & 90.8 & 88.9 & 89.1 & 88.4 & 93.8 & 90.2 & 93.1 & 88.8 & 88.8 & 92.6 & 93.5 & \textbf{97.4} & 89.1 & 92.8 & \underline{95.5}\\
		
		& k-NN(k=5) & 86.1 & 87.3 & 87.4 & 86.7 & 93.1 & 89.0 & 92.2 & 85.6 & 87.2 & 92.2 & 92.6 & \textbf{97.3} & 88.3 & 91.8 & \underline{95.3}\\ \hline
		
		\multirow{3}{*}{Faces94 PART 1} & k-NN(k=1) & 97.1 & 96.9 & 97.1 & 96.9 & \underline{97.2} & 96.6 & \underline{97.2} & 96.9 & 96.9 & 97.1 & 81.6 & \underline{97.2} & 96.5 & 97.0 & \textbf{97.3}\\
		
		& k-NN(k=3) & 95.4 & 95.0 & 95.4 & 95.2 & 95.3 & 94.6 & 93.9 & \underline{96.1} & 95.6 & \textbf{96.2} & 80.5 & 91.5 & 94.9 & 95.2 & \underline{96.1}\\
		
		& k-NN(k=5) & 94.6 & 94.6 & 94.7 & 93.8 & 95.3 & 92.9 & 93.6 & 94.0 & 94.4 & \textbf{96.2} & 81.4 & 89.2 & 93.4 & 93.8 & \underline{96.0}\\
		\hline
		\hline
	\end{tabular}
	\label{table4}
\end{table*}

\begin{table*}[!t]
	\renewcommand{\arraystretch}{1.2}
	\caption{Classification results (20\% labeled data) using k-NN classification algorithms. The best results are marked by bold font and the second best results are marked by underline. }
	\centering
	\setlength{\tabcolsep}{1mm}
	\begin{tabular}{c|c|ccccccccccccccc}
		\hline
		\hline
		Algorithms & Metric & Original & PCA & gLPCA & NMF & GNMF & NTF & LRNTF & GLTD & NTD & GNTD & NTT & GNTT & TR & NTR & GNTR\\ \hline
		
		\multirow{3}{*}{ORL} & k-NN(k=1) & 74.7 & 74.7 & 69.7 & 75.3 & 78.7 & 73.4 & \underline{81.3} & 68.1 & 75.3 & 79.8 & 60.4 & 77.9 & 57.0 & 77.4 & \textbf{81.8}\\
		
		& k-NN(k=3) & 59.1 & 52.2 & 56.3 & 61.7 & 70.3 & 57.6 & \underline{73.1} & 55.0 & 60.3 & 72.6 & 42.3 & 68.2 & 37.9 & 63.8 & \textbf{74.1}\\
		
		& k-NN(k=5) & 46.3 & 43.4 & 49.7 & 52.0 & 49.8 & 48.7 & 45.3 & 49.4 & \underline{52.7} & 44.8 & 38.5 & 43.1 & 35.3 & \textbf{56.1} & 44.9\\ \hline
		
		\multirow{3}{*}{FEI PART 1} & k-NN(k=1) & 53.7 & 55.0 & 57.5 & 59.8 & 72.4 & 55.6 & 74.2 & 57.5 & 55.9 & 74.9 & 15.5 & 68.4 & 52.4 & \underline{79.2} & \textbf{80.9}\\
		
		& k-NN(k=3) & 41.2 & 46.3 & 45.3 & 46.4 & 57.6 & 42.5 & 55.0 & 41.3 & 48.4 & \textbf{60.4} & 9.7 & 45.9 & 39.8 & 53.1 & \underline{59.4}\\
		
		& k-NN(k=5) & 32.7 & 32.5 & 36.7 & 36.7 & 39.8 & 32.0 & 39.5 & 33.5 & 40.7 & 45.6 & 8.8 & 30.6 & 32.0 & 40.5 & \underline{44.1}\\ \hline
		
		\multirow{3}{*}{GT} & k-NN(k=1) & 57.2 & 48.5 & 46.2 & 51.0 & 51.2 & 51.3 & 56.3 & 46.5 & 51.5 & 46.3 & 42.2 & 49.7 & 37.4 & \textbf{59.5} & \underline{57.3}\\
		
		& k-NN(k=3) & 42.5 & 33.2 & 37.2 & 40.8 & 43.6 & 39.4 & \underline{48.6} & 41.0 & 39.8 & 40.2 & 37.3 & 43.3 & 27.0 & 47.9 & \textbf{48.8}\\
		
		& k-NN(k=5) & 42.7 & 33.0 & 39.2 & 41.5 & 42.2 & 39.8 & \underline{46.8} & 40.8 & 39.6 & 38.5 & 37.6 & 40.3 & 26.7 & 45.9 & \textbf{47.8}\\ \hline
		
		\multirow{3}{*}{COIL-100 PART 1} & k-NN(k=1) & 80.3 & 77.9 & 77.9 & 78.2 & 86.9 & 80.6 & 85.0 & 78.0 & 77.5 & 85.0 & 88.5 & \textbf{97.1} & 85.5 & 89.3 & \underline{96.9}\\
		
		& k-NN(k=3) & 78.9 & 76.1 & 76.1 & 76.5 & 84.9 & 78.7 & 84.4 & 76.5 & 76.0 & 83.2 & 86.4 & \textbf{96.4} & 82.5 & 87.3 & \underline{96.3}\\
		
		& k-NN(k=5) & 77.5 & 74.1 & 74.1 & 75.6 & 83.8 & 77.4 & 83.1 & 74.5 & 74.8 & 82.3 & 84.5 & \textbf{94.2} & 80.8 & 85.6 & \textbf{96.1}\\ \hline
		
		\multirow{3}{*}{Faces94 PART 1} & k-NN(k=1) & 96.0 & 94.3 & 94.4 & 95.5 & 95.6 & 95.0 & 94.4 & 95.2 & 95.4 & \textbf{96.2} & 94.0 & 94.4 & 95.2 & 95.9 & \textbf{96.2}\\
		
		& k-NN(k=3) & 94.6 & 95.1 & 95.2 & 94.3 & 95.9 & 93.6 & 94.5 & 94.2 & 94.8 & \textbf{96.9} & 75.2 & 90.6 & 93.9 & 94.3 & \underline{96.7}\\
		
		& k-NN(k=5) & 93.6 & 92.6 & 92.5 & 93.0 & 95.7 & 91.8 & 94.4 & 92.2 & 93.5 & \textbf{96.8} & 76.9 & 90.8 & 93.1 & 93.3 & \underline{96.3}\\
		\hline
		\hline
	\end{tabular}
	\label{table3}
\end{table*}

\begin{figure*}[!t]
	\subfigure[ ]{
		\includegraphics[width=0.23\textwidth]{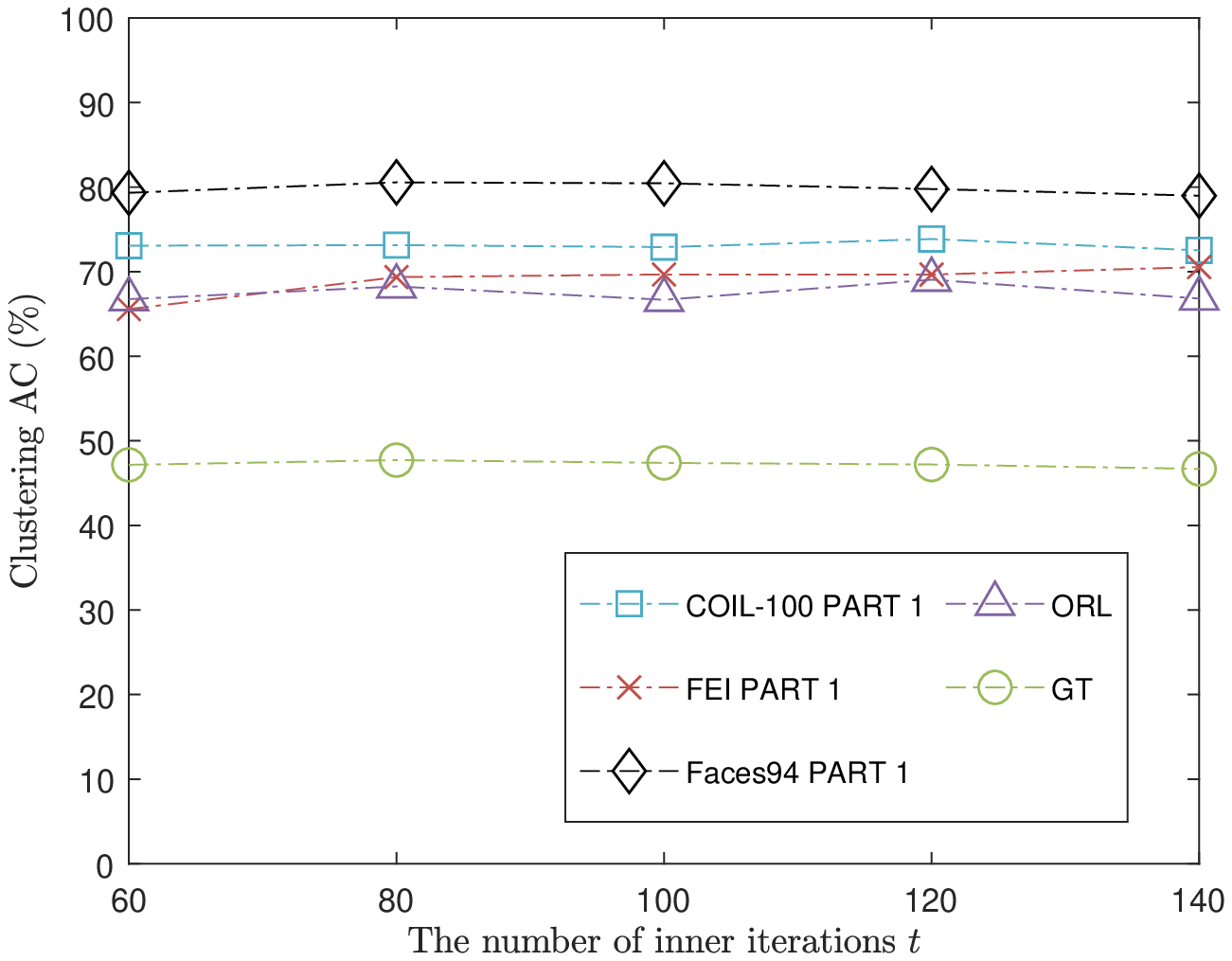}}
	\subfigure[ ]{
		\includegraphics[width=0.23\textwidth]{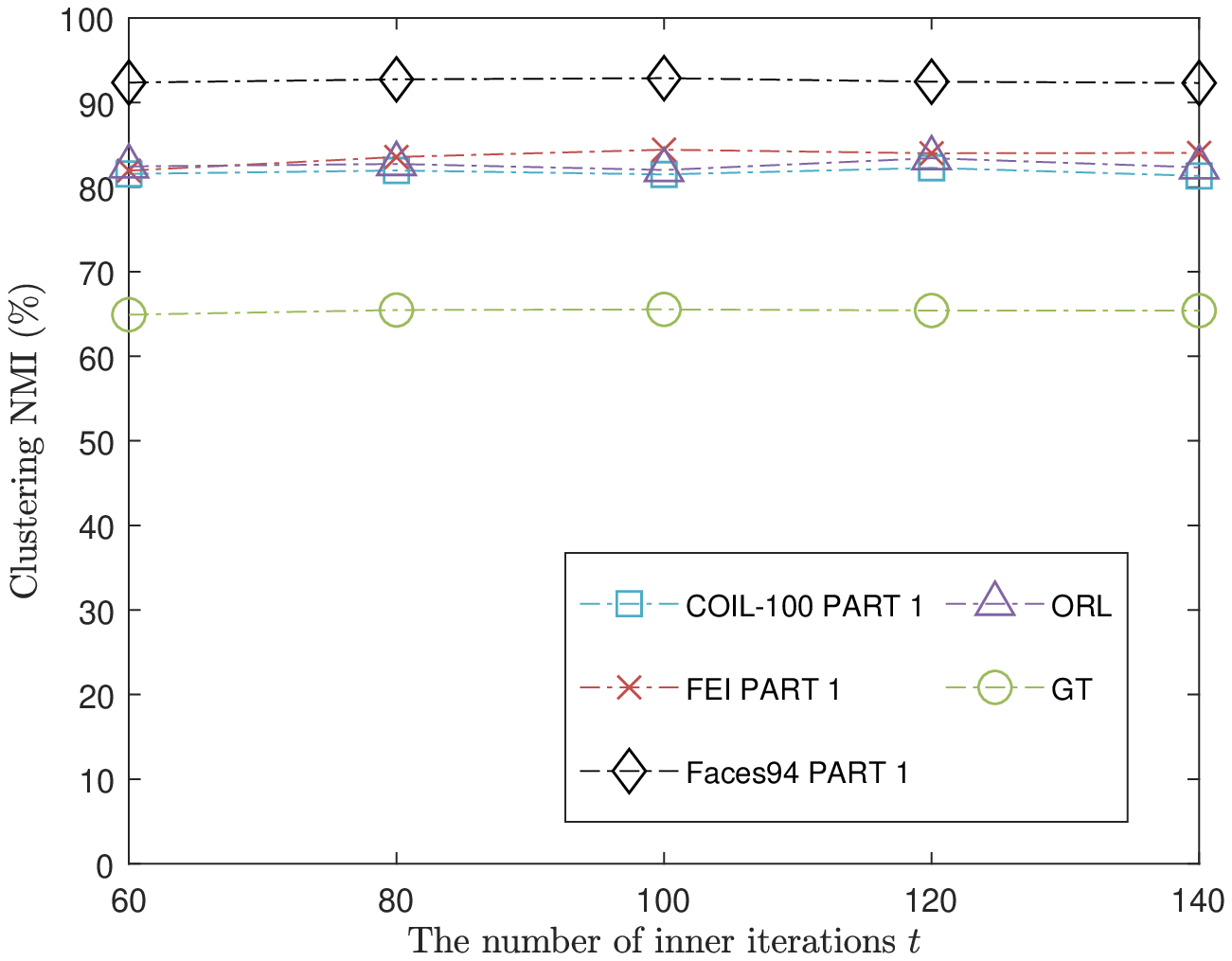}}
	\subfigure[ ]{
		\includegraphics[width=0.23\textwidth]{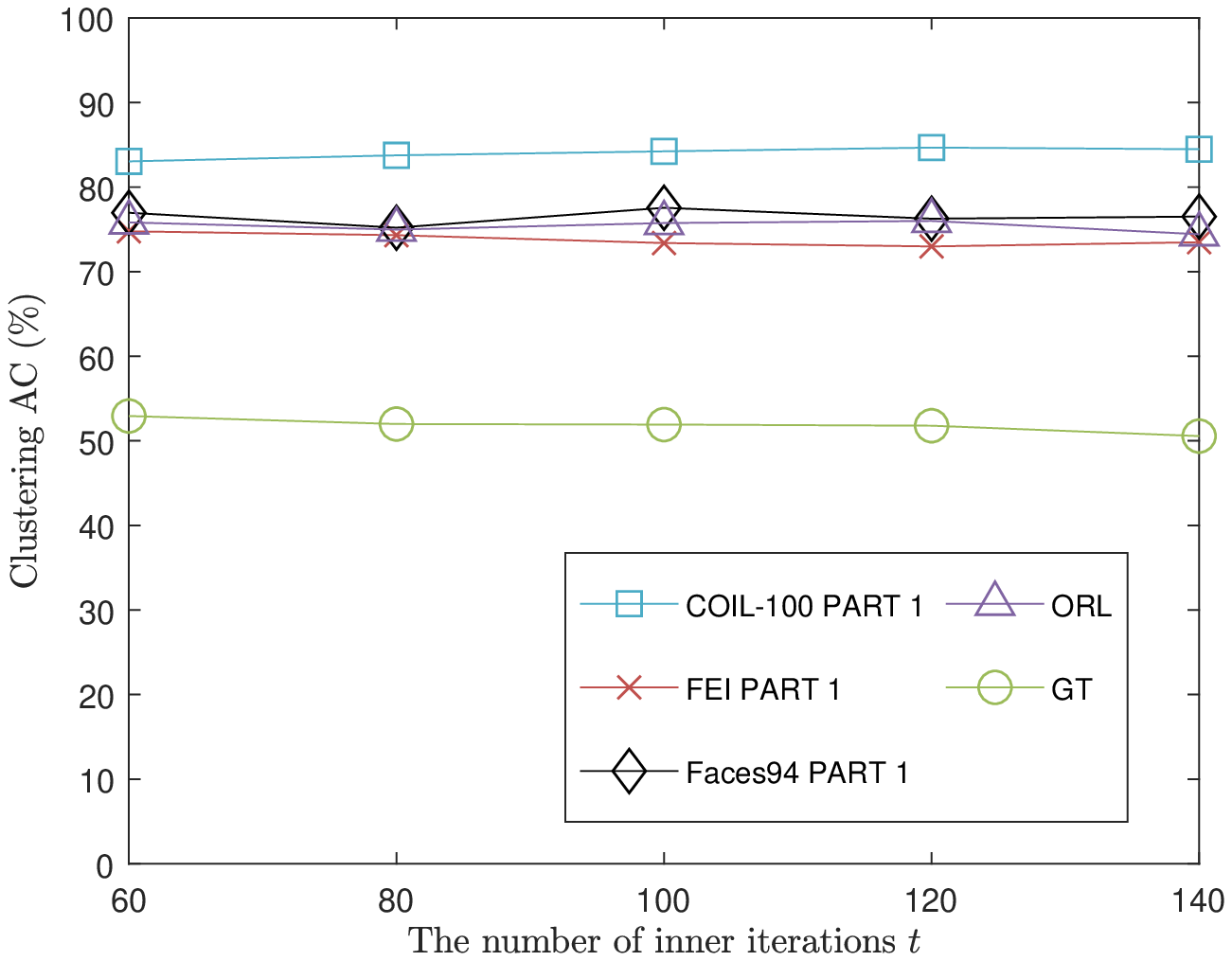}}
	\subfigure[ ]{
		\includegraphics[width=0.23\textwidth]{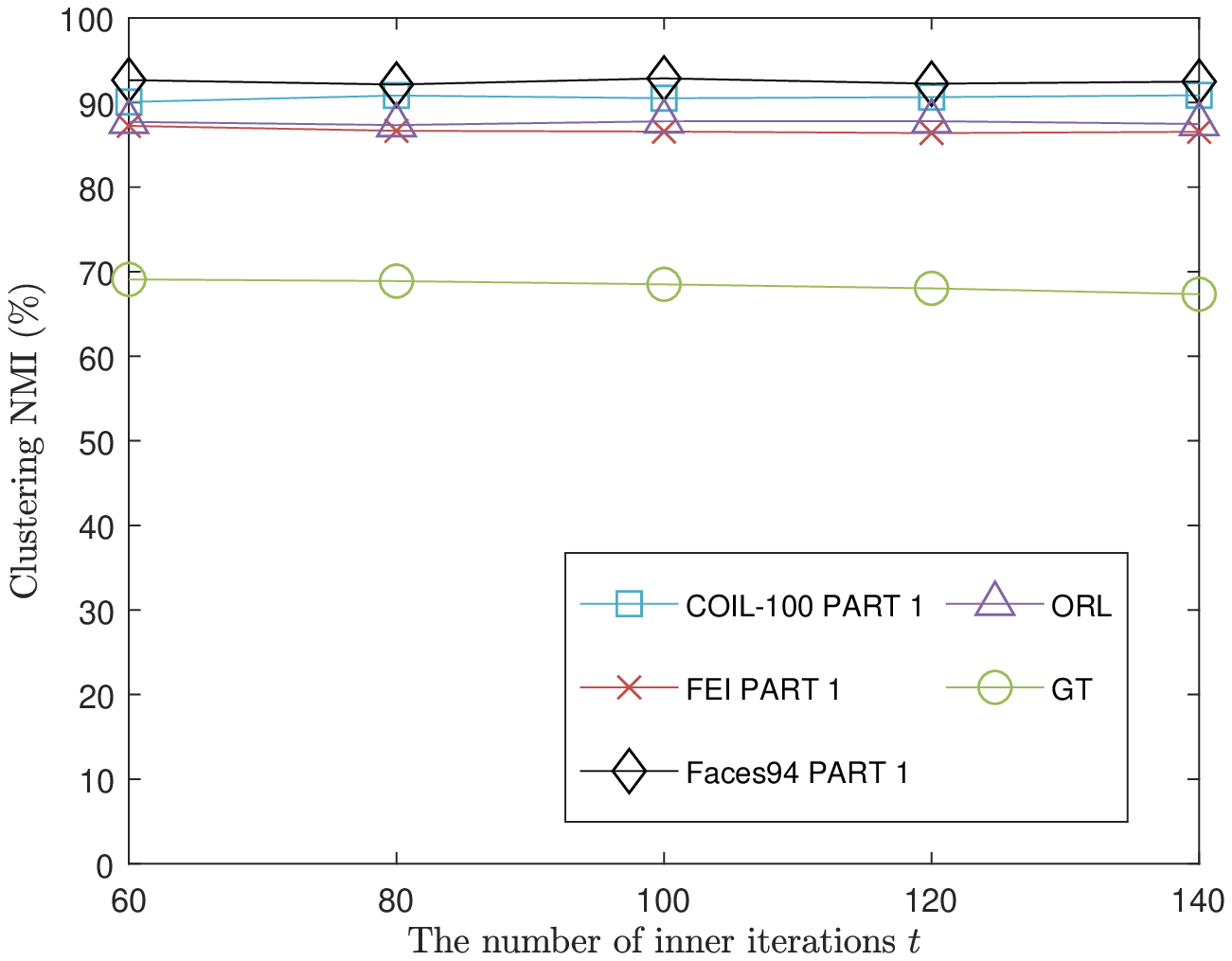}} 
	\\
	\subfigure[ ]{
		\includegraphics[width=0.23\textwidth]{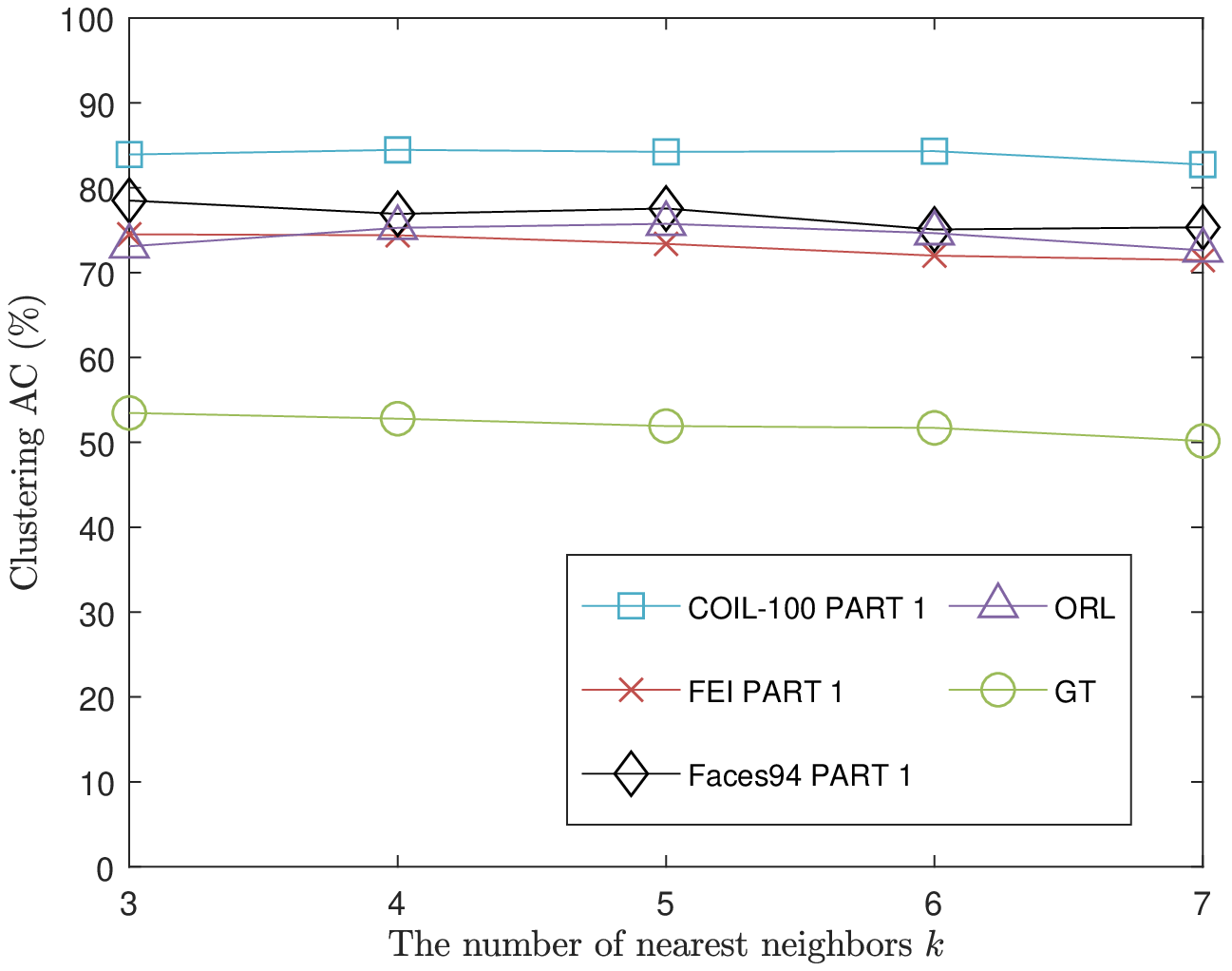}}
	\subfigure[ ]{
		\includegraphics[width=0.23\textwidth]{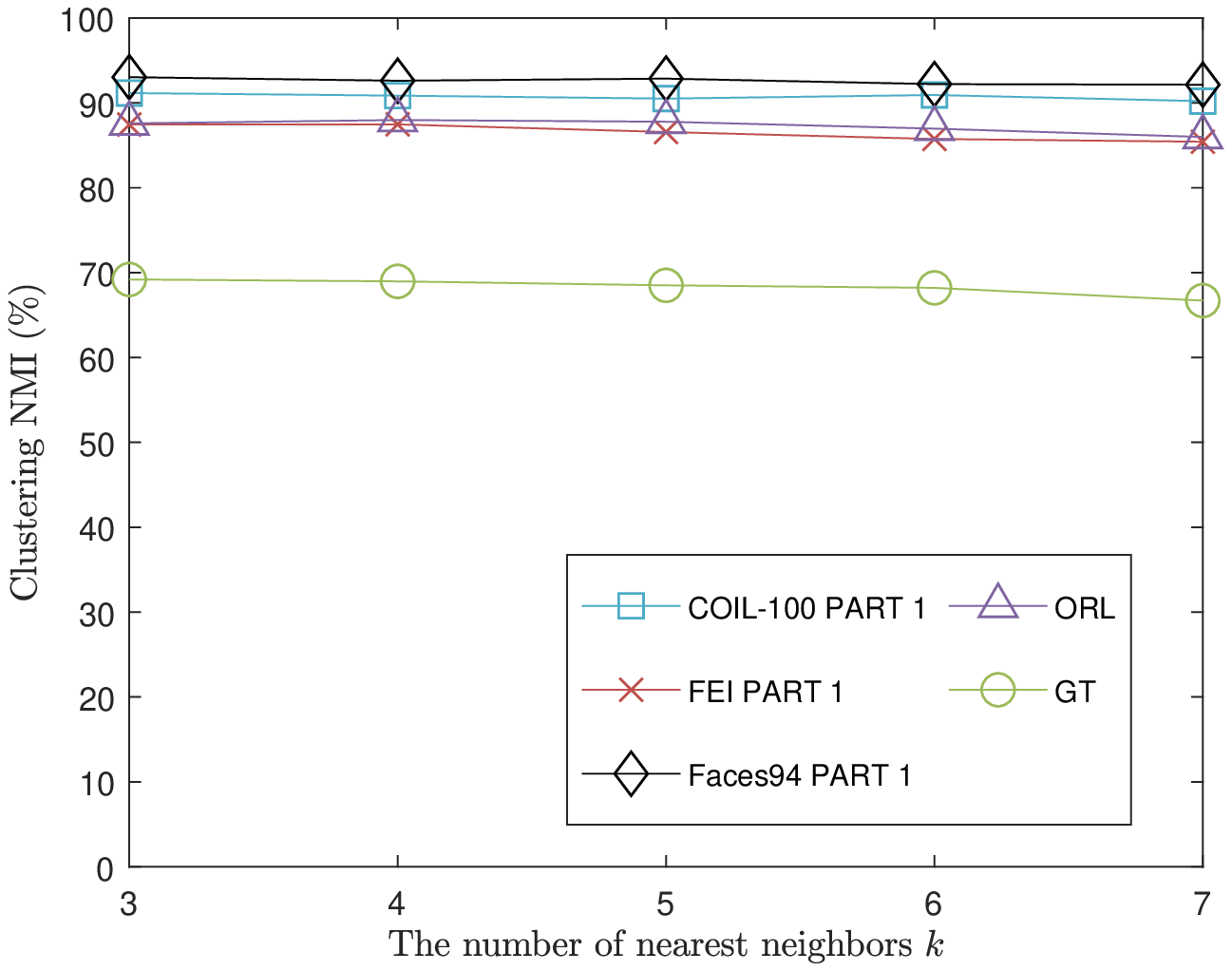}}
	\subfigure[ ]{
		\includegraphics[width=0.23\textwidth]{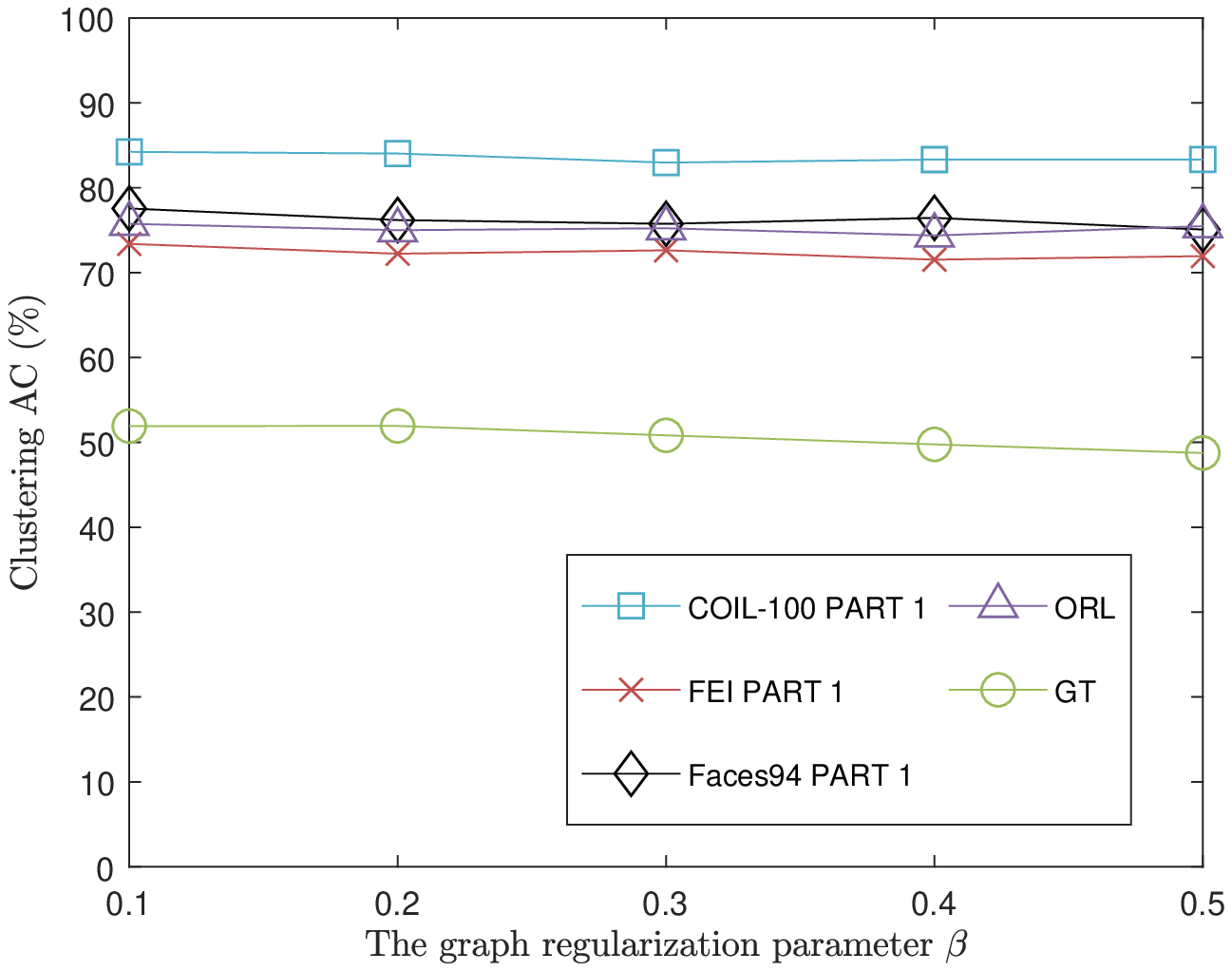}}
	\subfigure[ ]{
		\includegraphics[width=0.23\textwidth]{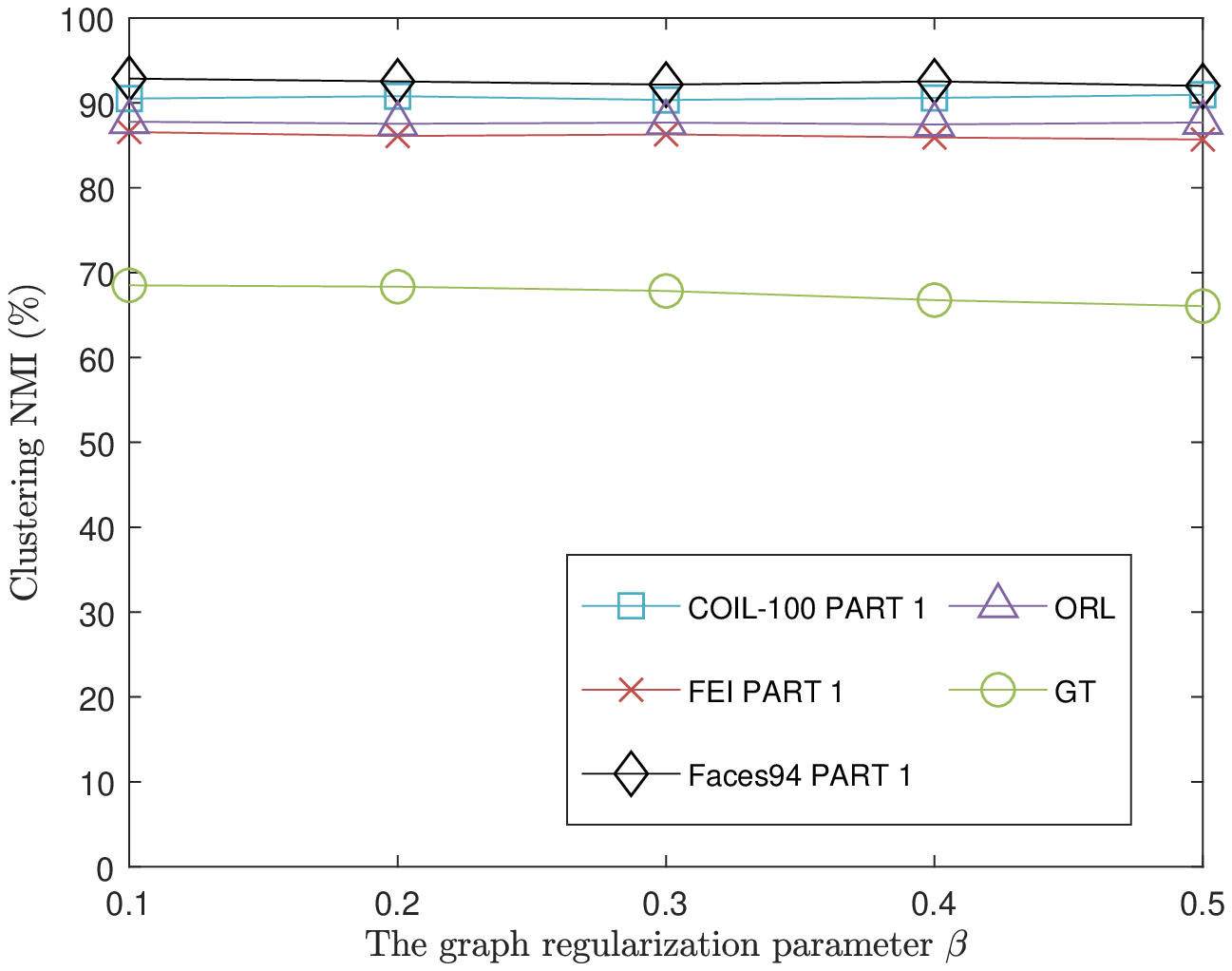}}
	\caption{Clustering performance of NTR and GNTR across different parameter $t$, $k$, $\alpha$. (a) and (b) AC and NMI of NTR across $60 \leq t \leq 140$. (c) and (d) AC and NMI of GNTR across $60 \leq t \leq 140$. (e) and (f) AC and NMI of GNTR across $3 \leq k \leq 7$. (g) and (h) AC and NMI of GNTR across $0.1 \leq \alpha \leq 0.5$}.
	\label{fig:6}
\end{figure*}

\begin{figure*}[!t]
	\subfigure[ ]{
		\includegraphics[width=0.185\textwidth]{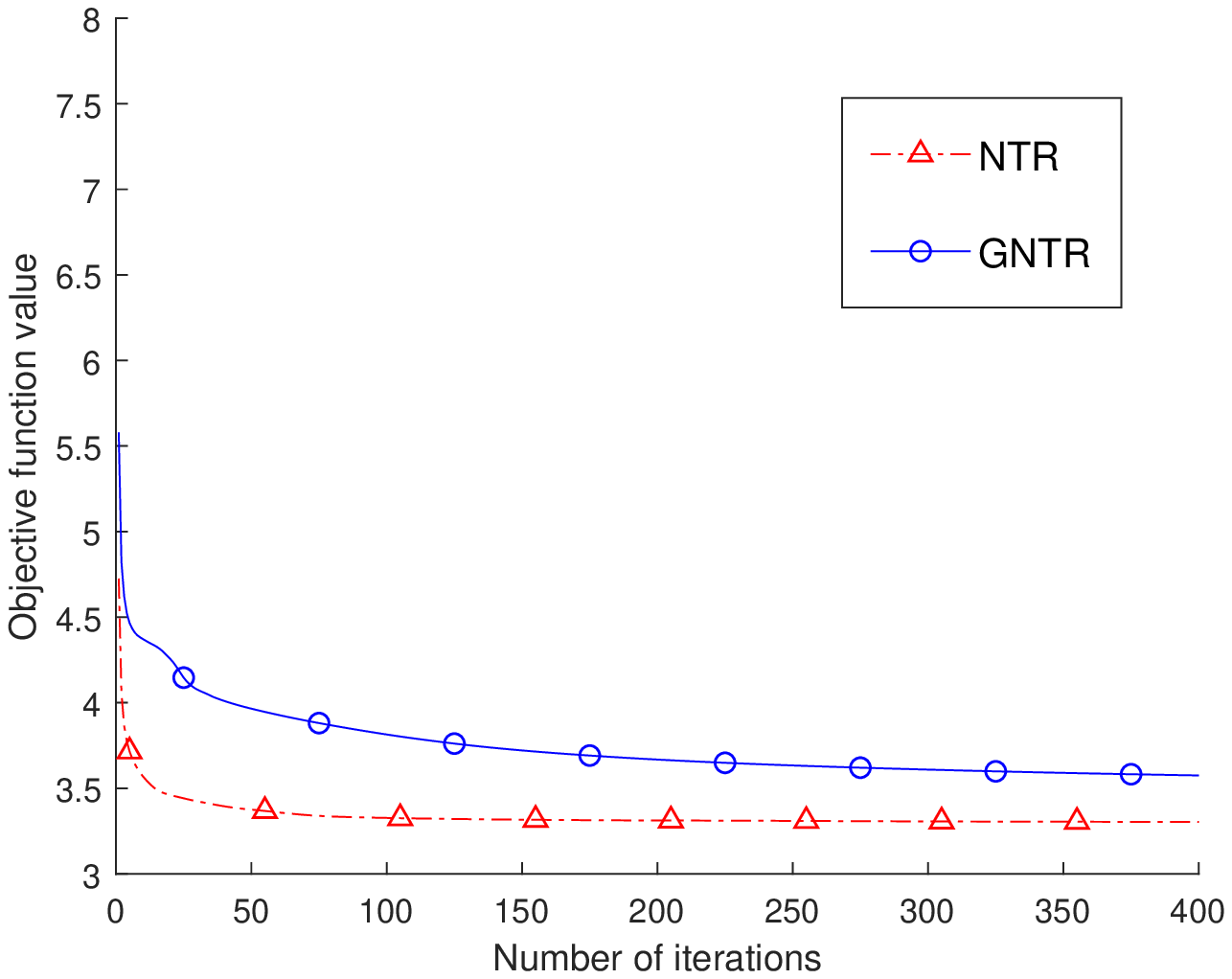}}
	\subfigure[ ]{
		\includegraphics[width=0.185\textwidth]{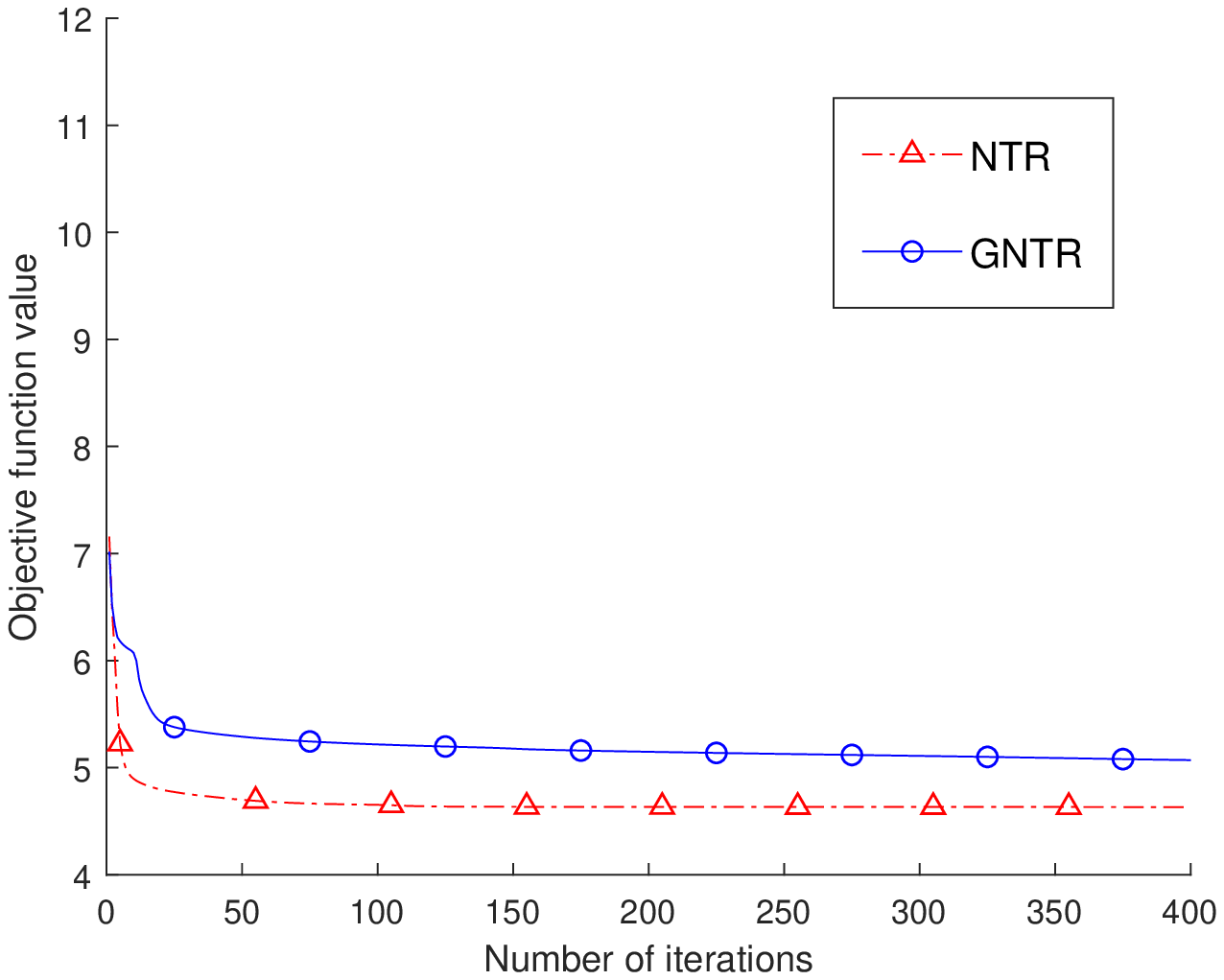}}
	\subfigure[ ]{
		\includegraphics[width=0.185\textwidth]{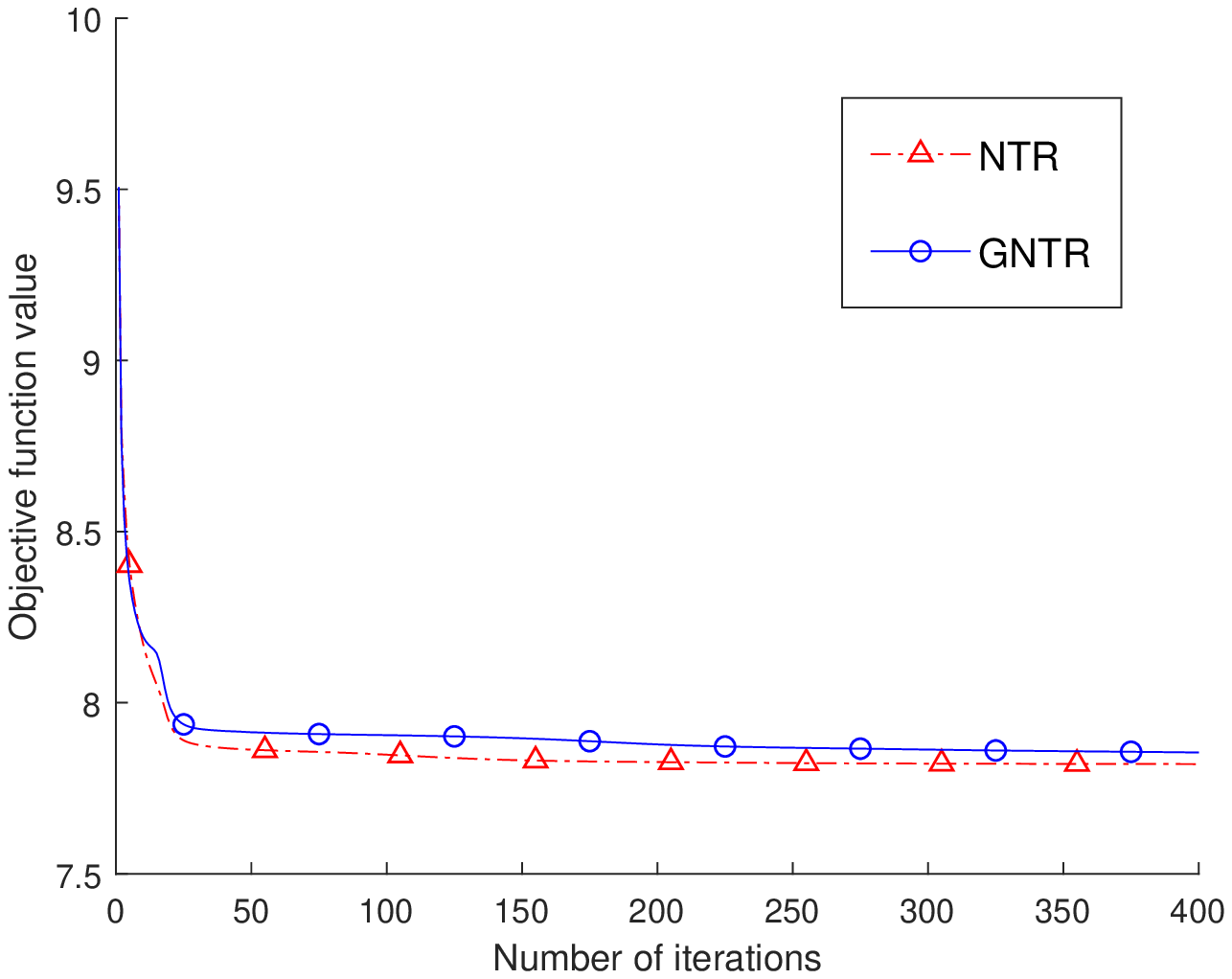}}
	\subfigure[ ]{
		\includegraphics[width=0.185\textwidth]{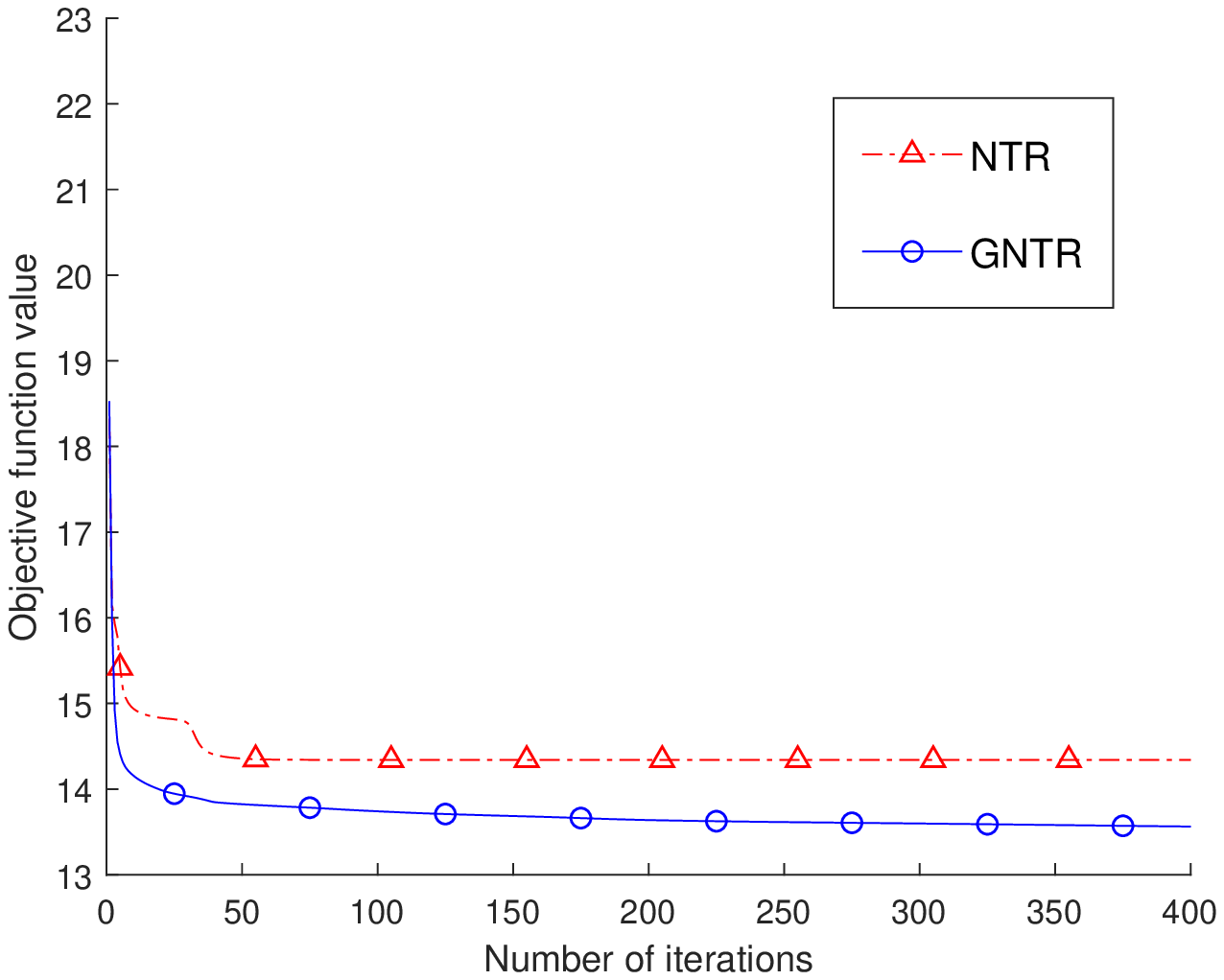}} 
	\subfigure[ ]{
		\includegraphics[width=0.185\textwidth]{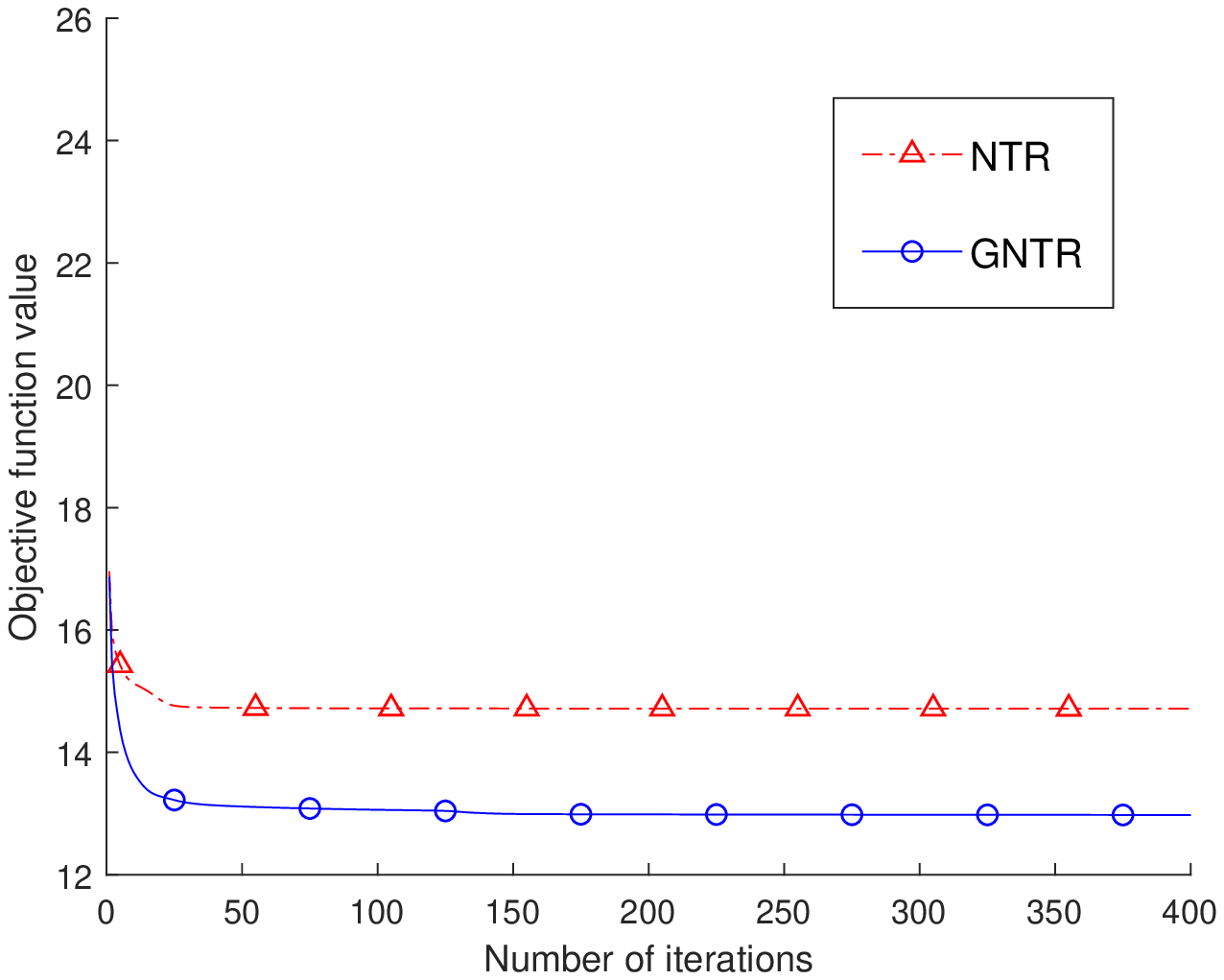}}
	\caption{Convergence curves of the proposed NTR algorithm and GNTR algorithm on five databases. (a) ORL database. (b) FEI PART 1 database. (c) GT database. (d) COIL-100 PART 1 database. (e) Faces94 PART 1 database.}
	\label{fig:7}
\end{figure*}

\subsection{Clustering and Classification tasks}
\label{sec:4.3}
To verify the performance our proposed algorithms, we compared them with the state-of-the-art algorithms in clustering and classification tasks.

\subsubsection{Clustering task}
\label{sec:4.3.1}
We perform the clustering analysis on the features of tensor objects extracted from each algorithm, and calculate the clustering evaluation measures to compare the proposed NTR and GNTR algorithms with the state-of-the-art algorithms

For simplicity, we uniformly set the number of the feature of tensor object to the number of categories $k$ of tensor data, which can achieve by adjusting the rank of algorithms. For Tucker-based algorithms, such as GLTD, NTD and GNTD, we empirically set the multiway rank as $r_{1}=r_{2}=10$, $r_{3}=3$, $r_{4}=k$ for the task of 4th-order database and set $r_{1}=r_{2}=10$, $r_{3}=k$ for the task of 3rd-order database. For the TT-based algorithms, such as NTT and GNTT, we set the nonnegative multiway rank to $2$ except for the multiway rank determined by category $k$ empirically. It's worth mentioning that the TR-based algorithms, such as TR, our proposed NTR and GNTR, decomposes the data into a series of 3rd-order core tensors, the features of the tensor objects is determined by two multiway rank of the feature core tensor. We set these two multiway rank as a pair of integer factors of $k$ and the other multiway rank are set as $2$ empirically. Most graph-based algorithms in addition to RPCA and GLTD, the regularization parameter $\beta$ is set to be $0.1$, respectively. For two graph-based algorithms RPCA and GLTD with orthogonal constraints, the $\beta$ is set to be $0.01$ empirically. For all the graph-based algorithms, the number of nearest neighbors $k$ is set to be $5$. To mitigate the local convergence issue, we repeat $200$ times with random initialization in each run of K-means. For each database, we repeat the above process of the task for $10$ times and report the average performance in Table \ref{table2}.

From the experimental results, we can draw the following conclusions:

\begin{itemize}
	\item The proposed NTR algorithm achieve better performance than the other nonnegative algorithms without graph regularized term. The proposed GNTR is superior to all other algorithms in most cases. Because GNTR not only inherits the advantages of NTR, but also learns the manifold geometry information of the data to further improves the recognition rate in clustering tasks.
	\item Compared with TR, NTR achieve better performance in most cases, which demonstrates limiting the data representation as nonnegative can makes the data representation more discriminative.
	\item The graph-based algorithms, such as GNMF, LRNTF and GNTD, these algorithms achieve better performance than the corresponding algorithms without graph regularization, such as NMF, NTF and NTD. This phenomenon can also be observed in \cite{wang2011image,qiu2019graph,jiang2018image}, and it suggests the importance of learn the manifold geometric information of tensor data.
\end{itemize}

\subsubsection{Classification task}
\label{sec:4.3.2}

To further analyze the effectiveness of our proposed algorithms in the data representation of tensor data, we perform classification tasks in the features extracted by each algorithms using k-NN classification algorithms. We respectively select first $20\%$ and $40\%$ of objects for each class as labeled data, and use the rest as unlabeled data in each task. We use the same comparison algorithm as in Section (\ref{sec:3.3}) to compare with the algorithms we proposed. We train the classifier on the features of labeled data and predict class labels on the features of unlabeled data. For each database, we repeat the above task for $10$ times and report the average classification accuracy. For the parameters of each algorithms, our settings are the same as Section \ref{sec:4.3.1}.

From Table \ref{table3} and Table \ref{table4}, it can be observed that the experimental results are similar to the clustering task results. The proposed algorithms NTR and GNTR perform the best in the most cases, which is further proving the advantages of NTR and GNTR algorithms in the multiway representation of tensor data.

It is worth noting that the NTT algorithm is achieve well performance in the task of COIL-100 PART 1 database, and conversely, achieve poor performance in the task of FEI PART 1 database. The probable reason is that the nonnegative multiway products of NTT core tensors must follow a strict order such that the optimized NTT core tensors highly depend on the permutation of tensor dimensions \cite{zhao2016tensor}. Therefore, the feature core tensor extracted of NTT can only maintain the direct connection and interaction with the penultimate core tensor that store the color information. The large color difference of different objects in the COIL-100 PART 1 database, which thus the NTT algorithm can achieve well performance in the task of this database. NTT achieve poor performance in the task of the FEI PART 1 database due to the lines of different face objects in this data set have large differences, not color differences. The performance of our proposed NTR and GNTR algorithms are better than NTT and GNTT algorithms in most cases, respectively. A probable reason is that our algorithms inherited the circular dimensional permutation invariance from TR decomposition \cite{zhao2016tensor}, the feature core tensor extracted of NTR and GNTR can maintain the direct connection and interaction with the first core tensor. Therefore, NTR and GNTR can comprehensively take into account the similarities and differences of the lines and colors of the tensor target to achieve better performance of the classification tasks.

\subsection{Parameter Selection}
\label{sec:4.4}
To investigate the parameter sensitivity of our proposed NTR and GNTR algorithms, we evaluated the effect on the performance clustering task of five databases by different parameters of two algorithms. NTR has fewer parameters, only the number of inner iterations $t$. For the GNTR algorithm, there are three parameters need to be predefined, which are the number of inner iterations $t$, the number of nearest neighbors $k$ and the graph regularization parameter $\beta$. We report the average clustering performance of NTR algorithm in the setting $t \in \{60,80,100,120,140\}$. For the GNTR algorithm, we report the average clustering performance in different settings:$\left(1\right)$ fix $k=5$ and $\beta = 0.1$, and choose $t \in \{60,80,100,120,140\}$; $\left(2\right)$ fix $t=100$ and $\beta =0.1$, and choose $k$ to vary from $3$ to $7$; $\left(3\right)$ fix $t=100$ and $k=5$, and choose $\beta \in \{0.1 0.2 0.3 0.4 0.5\}$.

Fig. \ref{fig:6}. presents the clustering performance of the NTR and GNTR across different parameter on the five public databases. It can be observed that the performances of NTR and GNTR algorithms across different parameters are quite stable. The AC and NMI of NTR and GNTR changes little when the number of inner iterations $t$ rises from $60$ to $140$, so $t$ can be selected around $100$ to balance the calculation cost and fitting error. It can be observed that the AC and NMI of NTR and GNTR are both stable when $k$ in range $3-7$. This phenomenon has also been observed in the paper \cite{qiu2020generalized}, $k$ has a relatively weaker influence on the performance of the algorithm. Therefore, the number of nearest neighbors $k$ is set in range $5$ in our clustering and classification tasks. It can be shown that the graph
regularization parameter $\beta$ also has a relatively weak effect
on the performance of the GNTR algorithm, and it is stable in range $0.1-0.5$.

\subsection{Convergence Study}
\label{sec:4.5}
In this section, we show the convergence curves of the proposed NTR algorithm and GNTR algorithm on five databases in Fig. \ref{fig:7}. The experimental results demonstrated that the proposed algorithms converges very quickly and usually taking less than 150 iterations.

\section{Conclusion}
\label{sec:5}
In this paper, we propose the NTR, which expresses the information of each dimension of tensor data by corresponding 3rd-order core tensors. The NTR can extract the parts-based basis with rich colors and rich lines of tensor objects, which can provide more interpretable and meaningful representation for physical signals. We also combine the graph regularization with NTR to develop GNTR, which perfectly inherits the advantages of NTR and enables the extracted data representation to preserve the manifold geometry information for tensor data. An efficient method based on accelerated proximate gradient method is developed to optimize our proposed algorithms, and it has been proved its convergence and efficiency. The experimental results demonstrated the effectiveness of our proposed algorithm. The parts-based basis extracted of our algorithms is rich colors and rich lines that provide more interpretable and meaningful representation for physical signals. The proposed GNTR algorithm can achieve better performance than state-of-the-art algorithms on clustering and classification tasks. In the future, we hope to combine the proposed algorithm with low-rank approximation technology to enhance the robustness and computational efficiency.

\section*{Acknowledgment}

The authors would like to thank...

\ifCLASSOPTIONcaptionsoff
  \newpage
\fi



%

\bibliographystyle{IEEEtran}

\bibliography{GNTR}

%








\end{document}